%% file: acl_latex.tex
\pdfoutput=1

\documentclass[11pt]{article}

\usepackage[]{acl}

\usepackage{times}
\usepackage{latexsym}

\usepackage{pgfplots}
\usepackage{pgfplotstable}
\pgfplotsset{compat=1.17}
\usepackage{tikz}
\usetikzlibrary{shapes,decorations,arrows,calc,arrows.meta,fit,positioning}
\usepackage{graphicx}
\usepackage{caption}
\usepackage{subcaption}
\usepackage{multirow}
\usepackage[percent]{overpic}

\usepackage{amsmath,amssymb}

\DeclareMathOperator{\EX}{\mathbb{E}}

\usepackage[export]{adjustbox}
\usepackage{rotating}
\usepackage{standalone}
\usepackage{booktabs, makecell, tabularx}
\newcolumntype{s}{>{\hsize=.05\hsize}X}
\newcommand{\heading}[1]{\multicolumn{1}{c}{#1}}

\usepackage[T1]{fontenc}

\usepackage[utf8]{inputenc}

\usepackage{microtype}

\title{When Does Syntax Mediate Neural Language Model Performance? Evidence from Dropout Probes}

\author{Mycal Tucker \\
  \texttt{mycal@mit.edu} \\\And
  Tiwalayo Eisape \\
  \texttt{eisape@mit.edu} \\\And
  Peng Qian \\
  \texttt{pqian@mit.edu} \\\AND
  Roger Levy \\
  \texttt{rplevy@mit.edu} \\\And
  Julie Shah \\
  \texttt{julie\_a\_shah@csail.mit.edu} \\}

\begin{document}
\maketitle
\begin{abstract}
Recent causal probing literature reveals when language models and syntactic probes use similar representations.
Such techniques may yield ``false negative'' causality results: models may use representations of syntax, but probes may have learned to use redundant encodings of the same syntactic information.
We demonstrate that models do encode syntactic information redundantly and introduce a new probe design that guides probes to consider all syntactic information present in embeddings.
Using these probes, we find evidence for the use of syntax in models where prior methods did not, allowing us to boost model performance by injecting syntactic information into representations.
\end{abstract}

\section{Introduction}
Recent large neural models like BERT and GPT-3 exhibit impressive performance on a large variety of linguistic tasks, from sentiment analysis to question-answering \citep{bertpaper,gpt3}.
Given the models' impressive performance, but also their complexity, researchers have developed tools to understand what patterns models have learned.
In probing literature, researchers develop ``probes:'' models designed to extract information from the representations of trained models \cite{linzen2016assessing,conneau-etal-2018-cram,probe-parser}.
For example, \citet{hewitt2019structural} demonstrated that one can train accurate linear classifiers to predict syntactic structure from BERT or ELMO embeddings.
These probes reveal what information is present in model embeddings but not how or if models use that information \citep{probingoverview}.

\begin{figure}[t]
    \centering
    \includegraphics[trim={0cm 0cm 0cm 0.0cm}, clip=true, width=\linewidth]{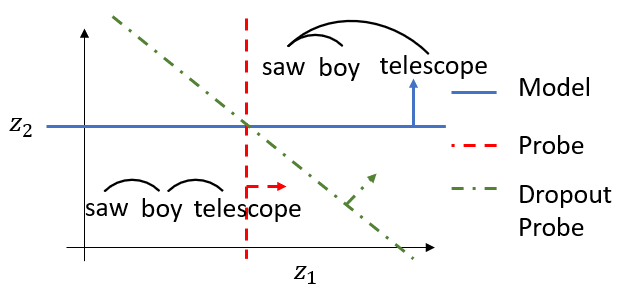}
    \caption{In a 2D embedding space, a model might redundantly encode syntactic representations of a sentence like ``the girl saw the boy with the telescope.'' Redundant encodings could cause misalignment between the model's decision boundary (blue) and a probe's (red). We introduce dropout probes (green) to use all informative dimensions.}
    \label{fig:simple}
\end{figure}

To address this gap, new research in causal analysis seeks to understand how aspects of models' representations affect their behavior \cite{elazar2020amnesic,insp,diagnostic,whatif,feder-etal:2021causaLM}.
Typically, these techniques create counterfactual representations that differ from the original according to some property (e.g., syntactic interpretation of the sentence). Researchers then compare outputs when using original and counterfactual embeddings to assess whether a property encoded in the representation is causally related to model behavior.

Unfortunately, negative results --- wherein researchers report that models do not appear to use a property causally --- are difficult to interpret.
Such failures can be attributed to a model truly not using the property (true negatives), or to a failure of the technique (false negatives). For example, as depicted in Figure~\ref{fig:simple}, if a language model encodes syntactic information redundantly (here illustrated in two-dimensions),
the model and probe may differentiate among parses along orthogonal dimensions.
When creating counterfactual representations with such probes, researchers could incorrectly conclude that the model does not use syntactic information.


In this work, we present new evidence for the causal use of syntactic representations on task performance in BERT, using newly-designed probes that take into account the potential redundancy in a model's internal representation.
First, we find evidence for representational redundancy in BERT-based models.
Based on these findings, we propose a new probe design that encourages the probe to use all relevant representations of syntax in model embeddings.
These probes are then used to assess if language models use representations of syntax causally, and, unlike prior art, we find that some fine-tuned models do exhibit signatures of causal use of syntactic information.
Lastly, having found that these models causally use representations of syntax, we used our probes to boost a question-answering model's performance by ``injecting'' syntactic information at test time.\footnote{Code at \url{https://github.com/mycal-tucker/mlm_dropout_probes}}

\section{Related Work}
\subsection{Language Model Probing}

Probing literature seeks to expose learned patterns of a neural language model by training small neural networks to map from model representations to human-interpretable properties \cite{Alain2017UnderstandingIL,conneau-etal-2018-cram,coenen2019visualizing}.
For example, \citet{hewitt2019structural} propose single-layered neural nets that map from embeddings to syntactic representations of sentences.
Such probing methods are correlative rather than causal because they depict what information is present in representations instead of how that information is used \cite{probe-parser,bookofwhy}.
Understanding when language models use structural information causally is an important question given the central role structure appears to play in human understanding of natural language \citep{chomsky1965aspects}.
In this work, we perform causal analysis by combining causal methods with a new probe design.

\subsection{Causal Analysis of Language Models}
Recently, researchers have begun applying causal analysis to language models to understand if and how they use human-interpretable properties in their decision making.
While direct text manipulations are sometimes possible (e.g., modifying ``The man works as a...'' to ``The woman works as a ...''), several methods rely on constructing counterfactual representations to measure model behavior \cite{Kaushik2020Learning,insp}.
Prior art has often found that standard models learn undesirable causal relationships by encoding unwanted biases or by not learning to rely upon syntactic principles \cite{feder-etal:2021causaLM,elazar2020amnesic}.

Our work is most closely related to \citet{whatif}, so we explain their technical approach here.
\citet{whatif} train non-linear structural probes (based on those designed by \citet{hewitt2019structural}) to predict aspects of a sentence's syntactic structure from model embeddings.
That is, a probe $p$ maps from an embedding, $z$, to a representation of syntax, $s$.
Trained probes are used to create counterfactual embeddings, $z'$, by updating $z'$ from $z$ via gradient descent to minimize a loss function, $L$, evaluated on the probe's output and a desired output based on an alternative syntactic interpretation, $s'$: $\nabla_{z'}L(p(z'), s')$.
Intuitively, these $z'$ are meant to represent ``what $z$ would have been if the structure of the sentence were $s'$.''
Using suites of syntactically ambiguous sentences, the authors measured how a model's outputs differed when using $z'$ generated from different syntactic interpretations.

While \citet{whatif} find that a pretrained BERT model does use representations of syntax causally (i.e., model outputs change when using different syntactic interpretations), the authors find that a BERT model fine-tuned on a question-answering task does \textit{not} show similar behavior.
Identifying causal mechanisms in models is important not only for fairness and robustness measures, but also for improving model performance.
In specific cases of subject-verb agreement, \citet{diagnostic} found that changing representations of a subject's plurality affected the plurality of verbs predicted by an LSTM.

In this work, we use the gradient-descent method proposed by \citet{whatif}, but we use a new probe design.
We identify several cases in which their method fails to uncover a causal relationship, whereas ours does.
Furthermore, compared to \citet{diagnostic}, we use more general representations of syntax instead of only plurality.

\begin{figure*}[t!]

    \begin{subfigure}[b]{0.48\textwidth}
        \centering
        \includegraphics[trim={0cm 0cm 0cm 0.0cm}, clip=true, width=0.98\textwidth]{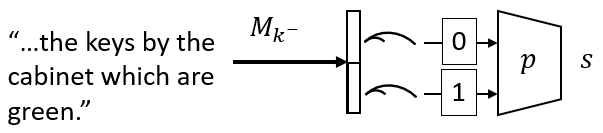}
        \caption{}
    \end{subfigure}
    ~
    \begin{subfigure}[b]{0.24\textwidth}
        \centering
        \includegraphics[trim={0cm 0cm 0cm 0.0cm}, clip=true, width=0.98\textwidth]{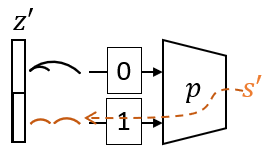}
        \caption{}
    \end{subfigure}
    ~
    \begin{subfigure}[b]{0.24\textwidth}
        \centering
        \includegraphics[trim={0cm 0cm 0cm 0.0cm}, clip=true, width=0.98\textwidth]{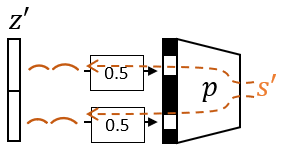}
        \caption{}
    \end{subfigure}
    
    \caption{If a model encodes the dependency structure of a sentence twice its embedding, a probe, $p$, may learn to ignore one copy of the information (indicated by learned weight 0) and only use the other (via learned weight 1) to predict $s$ (a). In such cases, the gradient of $s$ with respect to the embedding (dashed orange) only flows from one of the copies, so only that copy will be updated in counterfactual embeddings (b). However, by introducing a dropout layer that masks random inputs to the probe, dropout probes learn to use all informative parts of embeddings, which distributes the gradient across the whole embedding (c).}
    \label{fig:duplication}
\end{figure*}

\section{Technical Approach}
Here, we identify a limitation of prior causal probing art in which redundant information in embeddings could lead to probes and models using different representations of the same information, which in turn could lead to uninformative causal analysis results.
We propose a new probe architecture that addresses this limitation by encouraging probes to use all sources of information in embeddings.

\subsection{Limitations from Redundancy}
We show by example how prior art in causal probing may fail to reveal causal uses of syntactic information in language models.
Here, we use a simplified example; in later experiments we demonstrate that trained models exhibit similar phenomena.

In neural network probing literature, a probe, $p_\theta$, is a neural network parametrized by weights, $\theta$, that maps from representations, $z$, to a predicted property, $\hat{s}$: $\hat{s} = p_\theta(z)$.
\citet{hewitt2019structural} define two types of structural probes that map from $z$ to representations of a sentence's syntax.
The ``depth'' probe predicts words' depths in a parse tree; the ``distance'' probe predicts the distance between pairs of words in a parse tree.
In this paper, we assume $s$ refers to syntactic information, but probing techniques are general.
Given a corpus comprising $(z, s)$ pairs, probes are trained using supervised learning to minimize some loss.

Suppose that there exists a trained model, $M$; $M_{k-}$ (the first $k$ layers of $M$) encodes an input, $x$, into an embedding $z$.
The layers of $M$ after $k$, dubbed $M_{k+}$, produce a prediction, $\hat{y}$, from $z$.
For the purposes of this example, we state that $M$ uses syntactic information, and specifically that $z$ is informative of the syntactic structure of $x$.

Let us assume that the dependency structure of $x$ may be represented by within a vector, $z_{dep}$, and that $M_{k-}$ produces embeddings, $z$, which are two identical copies of $z_{dep}$.
Using pythonic notation, $z = [z_{dep}] + [z_{dep}]$.
Thus, $z$ contains syntactic information and, when we state that $M$ ``uses'' syntactic information, we formally mean that $\nabla_{z_{dep}} M_{k+}(z) \neq 0$.

Building upon this example, let us label the two copies of $z_{dep}$ as $z_{dep^1}$ and $z_{dep^2}$, although the two vectors remain identical.
If we train a probe to predict syntactic forms from $z$, it may arbitrarily learn to use any aspects of $z$ that are informative of its prediction, $s$.
Let us say that the probe learns to use only $z_{dep^2}$, again defined as $\nabla_{z_{dep^2}} p(z) \neq 0$.
However, $M_{k+}$ may only use $z_{dep^1}$: the copy that the probe does not use.


We claim that this example, while simplified, demonstrates a potential scenario in which causal probing techniques could return a false negative.
Specifically, if one generates counterfactual embeddings, $z'$, by changing $z$ according to the activations that change the probe's outputs, only $z_{dep^2}$ will change.
Because $M_{k+}$ uses only $z_{dep^1}$ for predictions, the model's output will not change.
This example is depicted in Figure~\ref{fig:duplication}.
Ultimately, without considering the redundancy in a model's internal representation, prior methods will fail to uncover the fact that $M$ actually does use representations of syntax causally.

\subsection{Dropout Probes}
In this section, we propose a neural probe architecture to address the limitations of prior art by encouraging probes to use all syntactic information present in $z$.
The desired behavior is depicted in Figure~\ref{fig:duplication}c: if the probe uses all activations that are informative of syntax, that will necessarily be a superset of the activations that the model uses for downstream processing (if the model uses syntax).
Therefore, when generating counterfactual embeddings using such probes, every activation encoding syntactic information would be updated, which in turn would change the model's output.

Our approach was inspired by an idea of creating a mixture of probes, each trained to use a different masked subset of activations in $z$.
The full set of such probes would have to learn to use all activations in $z$ that are informative of $s$.
One may approximate creating such a set by introducing a dropout layer as the first layer to a single probe.
At training time, the dropout layer masks a random subset of the input; the mask itself changes with every training batch.
We dub such probes ``dropout probes.''
This probe design, and our resulting findings when using them, are the main contributions of our work.
We note that adding a dropout layer to probes introduces a new hyperparameter but, in experiments, we found consistent results over a wide range of positive dropout values.

\section{Experiments}
Here, we report the results from three experiments establishing the benefits of dropout probes.
First, we found evidence supporting our hypothesis of redundantly-encoded syntactic information by calculating the mutual information between various activations in trained networks.
Second, we compared dropout probes to standard probes in a set of syntactically-ambiguous test domains.
We found that our method revealed evidence supporting the causal use of syntax in models where other methods did not \cite{whatif}.
Lastly, given our findings that models used syntax causally, we demonstrated how one could ``inject'' syntactic information into models to improve performance in syntactically-challenging tasks.

Experiments were conducted on four models, all based on huggingface's \texttt{bert-base-uncased} \cite{huggingface}.
The Mask model was the original model, trained on a masked language modeling task and next-sentence prediction \cite{bertpaper}.
The QA model was fine-tuned on the Stanford Question Answering Dataset 2.0 \cite{squad}.\footnote{The QA model was downloaded from huggingface model repository under ``twmkn9/bert-base-uncased-squad2''}
Lastly, we trained two models, dubbed NLI and NLI-HANS, that were finetuned on the Multi-Genre Natural Language Inference dataset or that dataset augmented with the Heuristic Analysis for NLI Systems (HANS) dataset, respectively \cite{mnli,rightwrong}.

The Mask model was used to compare our method to \citet{whatif}, who found that such models used syntactic information causally.
The QA model was used to study a finetuned model; prior art did not find evidence of causal use.
Lastly, the NLI models were used because Natural Language Inference is recognized as a difficult linguistic task that models appear to ``cheat'' on by leveraging spurious correlations in datasets \cite{rightwrong,nlifail1,nlifail2}.

\begin{table}[!tb]
    \centering
    \begin{tabular}{c|ccc}
         & $I(Z_1, D)$ & $I(Z_2, D)$ & $I(Z, D)$ \\
         \hline
         Mask & 2.2 & 2.6 & 2.7 \\
         QA   & 2.7 & 2.8 & 2.8 \\
         NLI  & 2.3 & 2.7 & 2.8 \\
    \end{tabular}
    \caption{The mean in nats of $I(Z, D)$ is less than $I(Z_1, D) + I(Z_2, D)$, indicating that information about $D$ is redundantly encoded in embeddings. Standard deviation under 0.2 for all values over 5 trials.}
    \label{tab:info}
\end{table}

\begin{figure*}[!tb]
    \setlength\tabcolsep{1pt}
    \settowidth\rotheadsize{Distance}
    \setkeys{Gin}{width=\hsize}
    \centering
    Mask Model Likelihood of Plural Candidates in Coordination Suite
    \begin{tabularx}{\linewidth}{sXX}
        & \heading{Dropout 0} & \heading{Dropout 0.4}\\
        \rothead{\centering Depth} &
            \includegraphics[trim={0.1cm 2.1cm 0.5cm 2.2cm}, clip=true, valign=m]{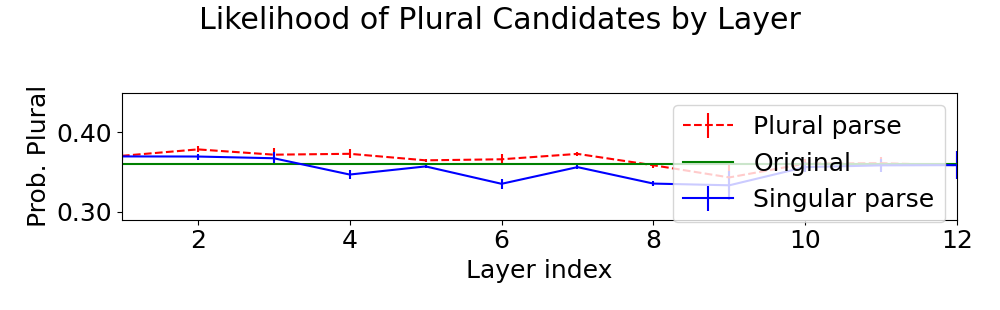} &
            \includegraphics[trim={0.1cm 2.1cm 0.5cm 2.2cm}, clip=true, valign=m]{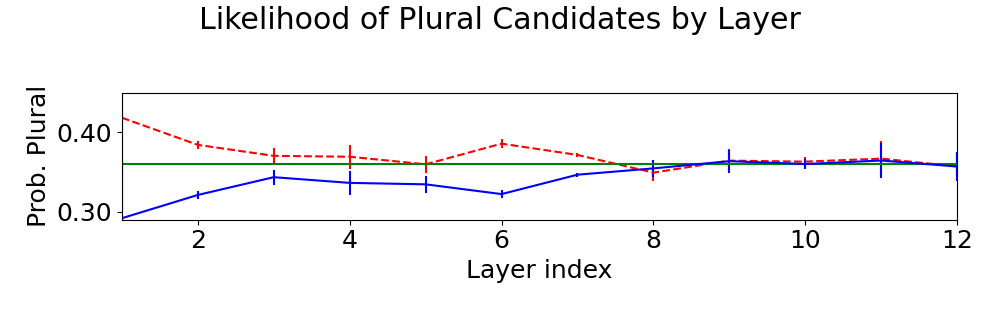}\\  
        \rothead{\centering Distance} &
            \includegraphics[trim={0.1cm 0cm 0.5cm 2.2cm}, clip=true, valign=m]{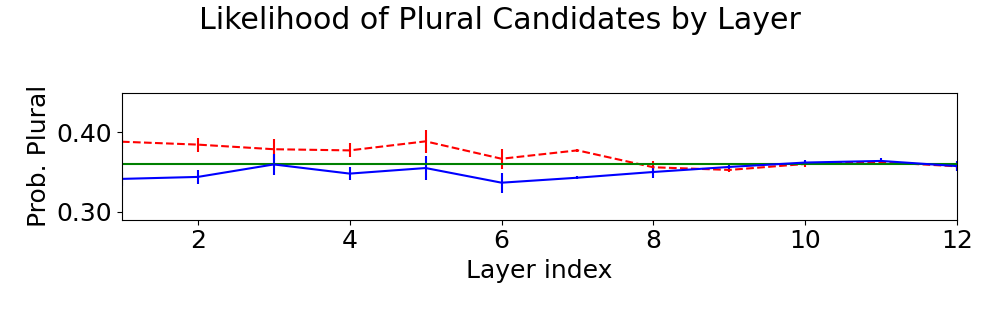} &
            \includegraphics[trim={0.1cm 0cm 0.5cm 2.2cm}, clip=true, valign=m]{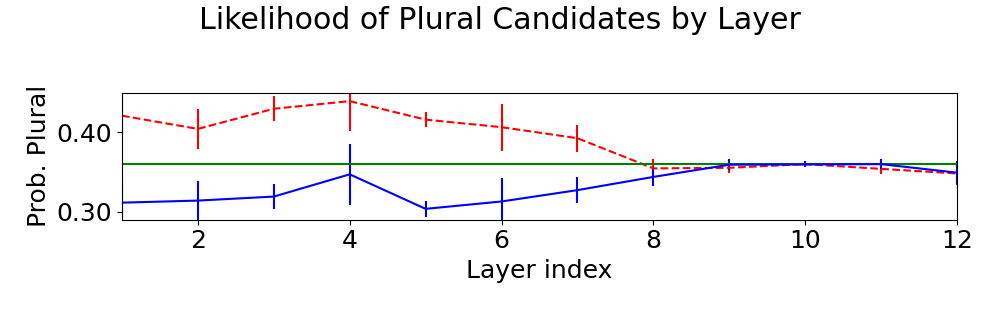}
    \end{tabularx}
        \caption{Mean and standard deviation probabilities over 5 trials for plural candidates using the original embeddings (green) or counterfactual embeddings favoring plural (dashed red) or singular (solid blue) parses. Counterfactual embeddings generated by both depth- and distance-based probes caused the greatest shift in model predictions.}
    \label{fig:mask_causal}
\end{figure*}

\subsection{Measuring Redundancy in Embeddings}
First, we found that language models redundantly encoded syntactic information in their embeddings, which motivated using dropout probes.

We used a technique from prior art, Mutual Information Neural Estimator (MINE), which is a neural-network based approach for estimating the mutual information between two random variables \cite{mine}.
It does so by computing a lower bound of mutual information and training a neural network to maximize that value.
This provides a conservative but tight estimate of mutual information.
We refer readers to Appendix~\ref{app:mine} for further details of our implementation.

We defined four random variables of interest.
The first, $D$, was the depth of each word in a sentence's parse tree; in other words, the labels used to train depth probes in prior literature \cite{hewitt2019structural}.
The second random variable, $Z$, was the 768-dimensional embeddings generated by a language model for each token in an input sentence.
Lastly, the third and fourth random variables ($Z_1$ and $Z_2$) corresponded to the first and second halves of $Z$ for each token.
That is, these variables comprised the starting and ending 384 units for each token's embedding.
By measuring the mutual information between different pairs of these variables, one may formalize our redundancy hypothesis into the following test: $I(Z, D) < I(Z_1, D) + I(Z_2, D)$.
Intuitively, if the test holds, there is shared syntactic information between $Z_1$ and $Z_2$.

We trained a MINE neural network on the first 5000 examples from the Penn TreeBank to estimate mutual information between random variables \cite{ptb}.
Embeddings were taken from the fourth layer of the MASK, QA, and NLI models, although they may be generated elsewhere.
Our results are presented in Table~\ref{tab:info}.
For all models, $I(D, Z) < I(D, Z_1) + I(D, Z_2)$; i.e., one gains little to no information for predicting $D$ from the full $Z$ instead of from just $Z_1$ or just $Z_2$.
This is evidence of redundant syntactic information in $Z$.

In these experiments using MINE, we demonstrated how $Z_1$ and $Z_2$ could be defined as the subsets of redundant activations depicted in Figure~\ref{fig:duplication}.
One could define other $Z_1$ and $Z_2$ to better characterize redundancy; here, we merely claim that at least some redundancy is present in the model embeddings.

\begin{figure*}[!htb]
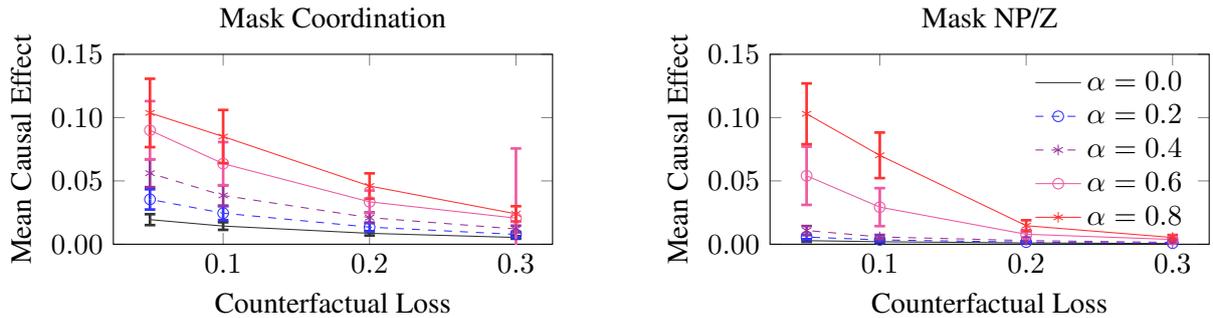

    \centering
    \begin{subfigure}[b]{0.46\textwidth}
        \centering
        \includestandalone[trim={0.12cm 0.0cm 0.0cm 0.0cm}, clip=true]{figures/tikz_diagrams/mask_coord_dist}
    \end{subfigure}
    \hfill
    \begin{subfigure}[b]{0.46\textwidth}
        \centering
        \includestandalone[trim={0.12cm 0.0cm 0.0cm 0.0cm}, clip=true]{figures/tikz_diagrams/mask_npz_dist}
    \end{subfigure}
    \caption{For the Coordination (left) and NP/Z (right) suites, interventions to a lower counterfactual loss ($x$ axis) and with higher-dropout probes (different curves) revealed the greatest causal effects. Means and standard errors.}
    \label{fig:mask_plots}
\end{figure*}

\subsection{Ambiguity Suite Experiments}
The prior section established that language models encode syntactic information redundantly; here, we showed that dropout probes overcame the challenges introduced by this redundancy by better aligning with models' true causal usage of syntax.
We compared dropout probes to the probes used in prior art via counterfactual experiments inspired by those used by \citet{whatif}.

We trained both distance- and depth-based probes, the two types of syntactic probes proposed by \citet{hewitt2019structural}.
We trained a new probe for each layer of each model, conducting 5 trials with random seeds 0 through 4.
All probes were implemented as 3-layer, non-linear neural nets that mapped from model embeddings (of dimension 768) through 2 ReLU layers of dimension 1024, to a final layer to predict a word's depth or distance in the parse tree from other words.
Probes were trained for up to 100 epochs, with early stopping based on validation set loss, using the Penn TreeBank dataset \cite{ptb}.
We found that this produced more accurate probes than prior art, which capped training at 30 epochs, and that these resulting probes did better than prior reported results, even without using dropout.
Each probe was prefixed by a dropout layer with a parameter, $\alpha$, that specified the proportion of inputs that were masked before being fed to the probe.
By setting $\alpha=0$, we recreated prior art of standard probes.
We additionally investigated positive values of $\alpha$ to measure the benefit of dropout.
Counterfactual embeddings were created via gradient descent through trained probes (with dropout disabled), as in prior art \cite{whatif}.
That is, new embeddings, $z'$, were generated to decrease the loss between $p(z')$ and a desired parse.
We called this loss the counterfactual loss.

In these experiments, we reported two types of results.
First, we visualized the effect of interventions, by layer, for a particular dropout rate and counterfactual loss.
This revealed that, typically, earlier layers in models were more susceptible to interventions.
Second, we devised an aggregate metric for the average difference, across all layers, in model outputs for counterfactuals generated with different parses.
This showed how lower counterfactual losses (i.e., more interventions) and higher dropout typically revealed larger effects.

Additionally, we note that the probes were trained to parse single sentences, but two of the models (QA and NLI) accepted two sentences as inputs.
For both models, counterfactual embeddings were creating by only updating the syntactically-ambiguous sentence and then concatenating it to the unaltered other embeddings.


\subsubsection{Masked Language Model}
In testing the Mask model, we largely reproduced patterns in prior results that such models use representations of syntax causally, although we found new results with dropout depth probes.
We tested the model with ambiguity test suites inspired by the Coordination and NP/Z suites from \citet{whatif}.
For example, in the Coordination suite, one sentence reads, ``The man saw the girl and the dog [MASK] tall.''
One may plausibly insert either a plural or singular noun in the masked location, depending upon the syntactic interpretation of the sentence.
We generated sentences using a template-based method; details of the prompts (and all prompts in this work) are included in Appendix~\ref{app:suites}.

The results of passing $z'$ generated from different parses in the Coordination suite through the rest of the Mask model are plotted in Figure~\ref{fig:mask_causal}.
The three plotted lines correspond to the model output using the normal embeddings (green), using $z'$ generated according to a parse favoring plural verbs (red dashed), or using $z'$ generated using parses implying singular verbs (blue solid).
The $y$ axis corresponds to the probability the model assigned to words implying a plural interpretation (``were,'' ``are,'' and ``as'') fitting in the masked location, normalized by the sum of probabilities assigned to those plural words or singular words (``was'' and ``is'').
If the Mask model uses syntactic representations correctly, counterfactuals from plural parses should increase the probability of plural words.

We indeed found that effect, although it is clearest when using dropout probes.
The causal effects using standard probes are plotted in the left column; we reproduced the findings from prior art that distance-based probes create the desired effect, but depth-based probes had little to no effect.
Conversely, when using dropout probes with $\alpha = 0.4$ (right column), we found much larger effects.

Averaging across all layers, we also measured the mean difference in output when using counterfactual embeddings generated according to different parses.
Intuitively, this generated a single number that captured the average difference between the red and blue lines in the plots in Figure~\ref{fig:mask_causal}.

For a range of dropout values and counterfactual losses, we plotted the mean causal effect for the Coordination and NP/Z suites in Figure~\ref{fig:mask_plots}, using distance probes.
For a given counterfactual loss, using higher dropout probes produced larger effects.
In addition, lower counterfactual losses (corresponding to more gradient steps) induced greater effects.
These trends also held true for depth-based probes (Appendix~\ref{app:varying}).
Overall, using the Mask model, we recreated prior art and found new evidence that models also use a depth-based representation of syntax.

\begin{figure*}[!htb]
    \setlength\tabcolsep{1pt}
    \settowidth\rotheadsize{Distance}
    \setkeys{Gin}{width=\hsize}
    \centering
    QA Model Causal Effect on Coordination Suite by Layer
    \begin{tabularx}{\linewidth}{sXX}
        & \heading{Dropout 0} & \heading{Dropout 0.4}\\
        \rothead{\centering Depth} &
            \includegraphics[trim={0cm 2.05cm 0.2cm 0.0cm}, clip=true, valign=m]{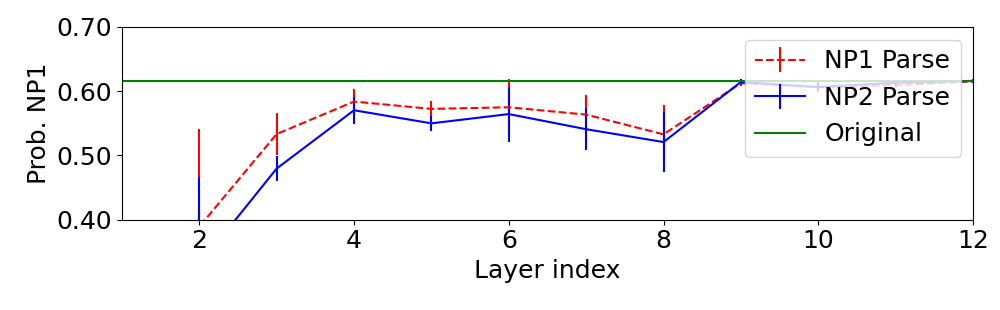} &
            \includegraphics[trim={0cm 2.05cm 0.2cm 0.0cm}, clip=true, valign=m]{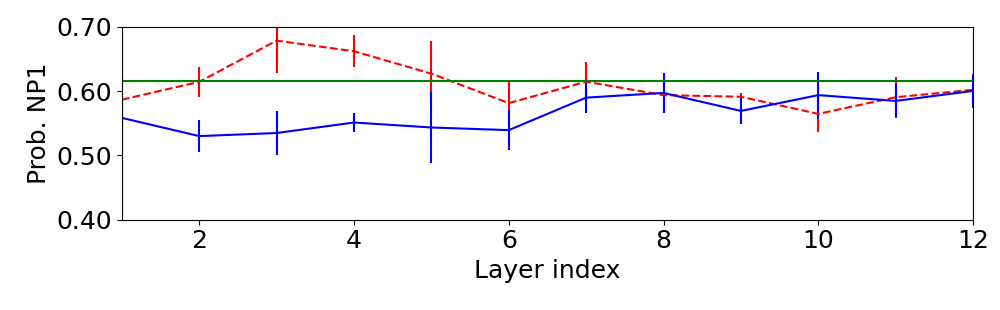}\\  
        \addlinespace[1pt]
        \rothead{\centering Distance} &
            \includegraphics[trim={0cm 0.4cm 0.2cm 0.0cm}, clip=true, valign=m]{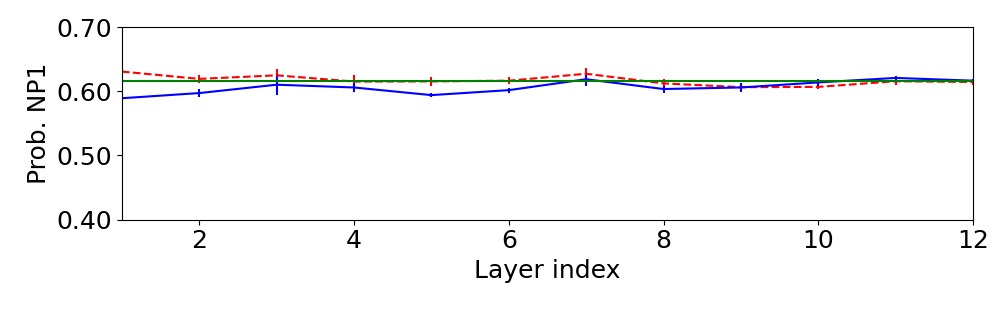} &
            \includegraphics[trim={0cm 0.4cm 0.2cm 0.0cm}, clip=true, valign=m]{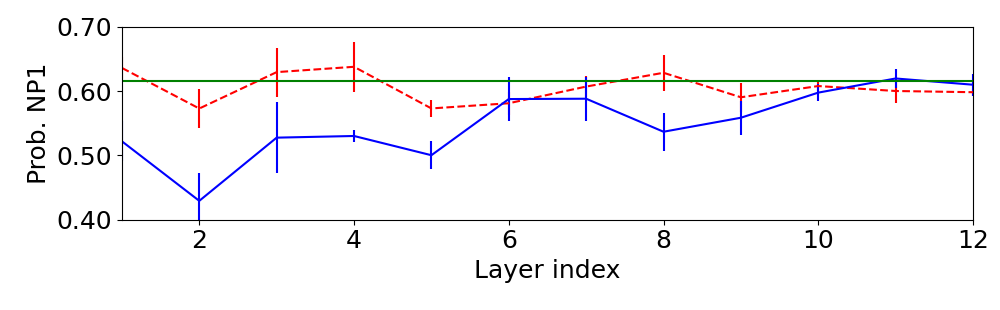}
    \end{tabularx}
        \caption{Causal effects for the QA model using depth- (top row) and distance- (bottom row) based probes with dropout of 0 (left column) or 0.4 (right column) on the Coordination corpus to counterfactual loss 0.05. Dropout probes produce more stable and larger effect sizes. Means and standard deviations over 5 trials plotted.}
    \label{fig:qa_causal_coord}
\end{figure*}


\setlength{\fboxsep}{0pt}

\begin{figure*}[!htb]
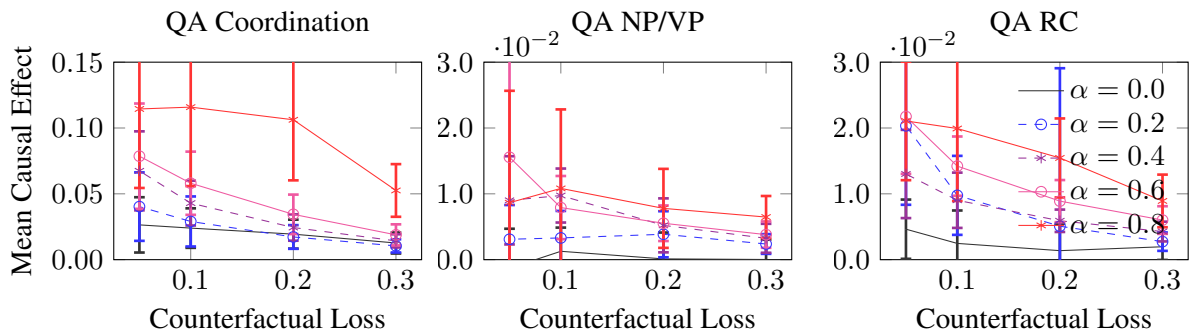

    \centering
    \begin{subfigure}[b]{0.32\textwidth}
        \includestandalone[trim={0.12cm 0.0cm 0.0cm 0.0cm}, clip=true]{figures/tikz_diagrams/qa_coord_depth}
    \end{subfigure}
    ~
    \begin{subfigure}[b]{0.32\textwidth}
        \centering
        \includestandalone[trim={0.25cm 0.0cm 0.0cm 0.0cm}, clip=true]{figures/tikz_diagrams/qa_npvp_depth}
    \end{subfigure}
    \begin{subfigure}[b]{0.32\textwidth}
        \centering
        \includestandalone[trim={0.25cm 0.0cm 0.0cm 0.0cm}, clip=true]{figures/tikz_diagrams/qa_rc_depth}
    \end{subfigure}
    \caption{Mean causal effects when using depth-based probes for the QA test suites. Smaller counterfactual losses and higher dropout rates typically induced greater effects, although the scale of the effects varied by suite (note different axis scales). Means and standard deviations over 5 trials.}
    \label{fig:qa_plots}
\end{figure*}

\subsubsection{QA Model}
We also found that the QA model used representations of syntax causally, contrary to prior findings, through a series of similar causal analysis experiments using syntactically-ambiguous inputs.
The QA model is a BERT-based model fine-tuned on a question-answering task to map from context and a question to a continuous span of the context that answered the question \cite{squad}.

We performed experiments using depth- and distance-based probes, using dropout values at increments of 0.1 from 0 to 0.9. 
We used three test suites for analyzing the causal use of syntax in the QA model: ``Coordination'', ``Relative Clause'' (RC), and a ``Noun Phrase/Verb Phrase'' (NP/VP) suite.
The Coordination suite consisted of 256 prompts with coordination ambiguity like, ``I saw the men and the women were tall. Who was tall?''
The RC suite consisted of 193 prompts with attachment ambiguity of a relative clause like, ``I saw the women and the men who were tall. Who was tall?''
The NP/VP suite consisted of 256 prompts like, ``The girl saw the boy with the telescope. Who had the telescope?''
Prompts were designed such that answers were dictated by syntactic interpretations.

Findings for the Coordination suite are plotted in Figure~\ref{fig:qa_causal_coord}.
On the $y$ axis, we plotted the model's prediction of words in the first noun phrase (NP1) starting the answer.
Correct causal use of representations of syntax would move the red line (corresponding to parses indicating NP1) above the original outputs, in green, and the blue line (for the other parse) below.

Unlike prior art, we found evidence that QA models use representations of syntax causally.
In the left column of Figure~\ref{fig:qa_causal_coord}, we found similar results to prior art: using standard depth-based probes produced noisy results, and distance-based probes had a small effect.
(In fact, this effect size shrank if we only trained the distance probe for 30 epochs, as in prior art, instead of the 100 epochs we used, indicating the importance of well-trained probes.)
In contrast to the standard probes, the dropout probes, plotted in the right column, revealed much larger effects of syntactic interventions.

More systematic analysis for all dropout rates, using distance and depth-based probes for all 3 test suites confirmed these trends.
We plotted the aggregate metrics for all suites using depth probes in Figure~\ref{fig:qa_plots}.
The causal effects were smaller in the RC and NP/VP suites than in the Coordination suite, indicating that the model may have learned a weaker causal link for these syntactic relations.
Nevertheless, all suites demonstrate the importance of using dropout in probes: without dropout (solid black curve), the causal effects were smaller than for any positive dropout rate.

We note briefly that the causal effects uncovered by dropout probes may not be solely attributed to dropout probes performing better at their parsing task.
In fact, adding dropout worsened probe performance according to typical probe performance metrics (Appendix~\ref{app:probe_perf}).

\subsubsection{NLI Model}
Lastly, we performed similar causal analysis on the NLI and NLI-HANS models and, in contrast to the Mask and QA models, we found no evidence for the causal use of syntax using any of our probes for either model.
The NLI model was finetuned on just the MNLI corpus, and the NLI-HANS model was finetuned with both the MNLI and HANS corpora, based on code from \citet{bert-nli-code}.
The NLI model had a test set accuracy of 86\%, and the NLI-HANS model had test set accuracy of 93\%.

We used a test suite based on the Coordination suite already introduced in this work: an example prompt was ``The person saw the keys in the cabinets which are green. The keys are green.''
The models had to classify these inputs among three classes of entailment, contradiction, or neutrality.



Ultimately, we failed to find any evidence that either the NLI or the NLI-HANS model used syntactic information causally.
The models always predicted entailment for all prompts, whether using original embeddings or counterfactuals generated for different parses.
We used distance probes with dropout values from 0 to 0.9 and created counterfactuals for losses from 0.05 to 0.3 and never observed a shift in predicted probability mass of more than 1\% when using counterfactuals.
Unfortunately, this suggests that simply augmenting the MNLI dataset with HANS may not be enough to produce a model that uses syntactic information causally.

\begin{figure}[!tb]
    \centering
    QA F1 via Interventions
    \begin{subfigure}[b]{0.48\textwidth}
        \includegraphics[trim={0cm 0.4cm 0cm 0.2cm}, clip=true, width=\textwidth]{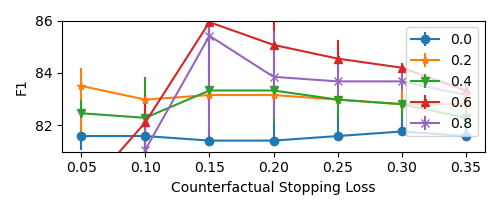}
    \end{subfigure}
    \caption{Using dropout probes over a range of dropout values (different curves) and counterfactual stopping losses improved model performance, and dropout typically improved performance. Medians and quartiles plotted over 5 trials.}
    \label{fig:oracle}
\end{figure}

\subsection{Boosting Performance with Probes}
\label{sec:boosting}
Earlier, we demonstrated that the QA model causally used representations of syntax for predictions; 
here, we showed that we could improve QA model performance at test time by ``injecting'' syntactic information into embeddings.
Because prior art had not found that QA models used syntax causally, such interventions were not previously pursued, as far as we are aware.


We designed a new, syntactically challenging ``Intervene'' test suite of 288 prompts for the QA model.
Example prompts are ``The person saw the keys by the cabinet which was green. What was green?'' and ``The person saw the keys by the cabinet which were green. What was green?''
Answering correctly (``the cabinet'' first and ``the keys'' second) depends upon using noun-verb agreement.
We used template-generated parse trees for each sentence and distance probes to create counterfactual embeddings for each sentence at layer 4 of the QA model.
Layer 4 was chosen based on performance on a validation dataset (Appendix~\ref{app:hypers}).

We passed the original and counterfactual embeddings through the QA model and measured performance on a test suite.
F1 performance is plotted in Figure~\ref{fig:oracle}; exact match metrics had similar trends.
Typically, higher-dropout probes improved performance more, although the highest-dropout probes deteriorated for the lowest counterfactual losses.
We hypothesize that this deterioration corresponded to generating out-of-distribution embeddings, but this topic warrants further study.

Lastly, we performed a similar experiment using the NLI and NLI-HANS models using 486 prompts drawn from the HANS dataset like ``The doctor near the actor danced. The actor danced'' \cite{rightwrong}.
The NLI model achieved 50\% accuracy (always predicting entailment) and the NLI-HANS model achieved 99\% accuracy.
Neither model's accuracy changed significantly when using counterfactuals with the correct parse for the first sentence, yet again indicating that these models may not use representations of syntax causally.




\section{Contributions and Conclusion}
In this work, we designed and evaluated ``dropout probes,'' a new neural probing architecture for generating useful causal analysis of trained language models.
Our technical contribution --- adding a dropout layer before probes --- was inspired by a theory of redundant syntactic encodings in models.
Our results fit within three categories: we showed that $1)$ models encoded syntactic information redundantly, $2)$ dropout probes, unlike standard probes, revealed that QA models used syntactic representations causally, and $3)$ by injecting syntactic information at test time in syntactically-challenging domains, we could increase model performance without retraining.

Despite our step towards better understanding of pretrained models, future work remains.
Natural extensions include studying pretrained models beyond those considered in this work, further research into redundancy in embeddings, more investigation into inserting symbolic knowledge into neural representations, and new methods for training models to respond appropriately to interventions.

\section{Ethical and Broader Impacts}
While the majority of this paper details the technical contributions of our work, here, we briefly consider some of the possible consequences of our findings based on transparency and causal modeling.

Fundamentally, we believe that causal analysis of models is a powerful tool towards more ethical AI.
Our dropout probes enable better inspection of models, providing possible mechanisms for regulators, ethicists, and even the general public to better understand AI systems with which they interact.
By injecting information into models at test time, as demonstrated in Section~\ref{sec:boosting}, we provide another mechanism for people to control model behavior.
Thus, our tool may reinforce values of transparency and value-alignment in AI, contingent upon access to the model for probing.

While we hope that our probing mechanism will be used for good, misuse of the tool is certainly possible.
In particular, the very causal rules that our tool uncovers may be used to reinforce biases.
For example, people may attempt to argue that a gender bias exhibited by a model are evidence of the ``correctness'' of that bias.
We urge readers to remember that models likely reflect biases present in human-generated data and certainly not ``true'' stereotypes.

We also note that the transparency benefits of our technique are not universally accessible.
Training a single probe on a single layer took approximately 2 minutes on an NVIDIA GeForce 3080; generating counterfactuals took approximately 1 second per counterfactul on similar hardware.
Although these operations individually are relatively lightweight (and certainly less computationally intensive than finetuning a whole model), systematic evaluation of models for many layers, multiple probes, and many counterfactuals is more challenging.
Furthermore, all analysis assumes access to the internals of the pretrained model itself.

Lastly, while our work is limited to diagnosis of existing models, we hope that it will enable important future research in causally-motivated models.
We hope to ultimately develop models that blend causal rules based on human guidance with emergent learned patterns from data.
Our work can complement such research by certifying that models have indeed learned the desired rules.

\section{Acknowledgements}
TE acknowledges support from the GEM consortium and the National Science Foundation Graduate Research
Fellowship under Grant No. 1745302.
\bibliography{anthology,custom}
\bibliographystyle{acl_natbib}

\appendix

\section{MINE Details}
\label{app:mine}
The Mutual Information Neural Estimator technique works by training a neural net to compute and maximize a lower bound on mutual information between two random variables \cite{mine}.
We describe the intuition of the technique, as well as our implementation, in this section; we refer readers to the full paper for theoretical analysis of MINE.

The mutual information between two variables is defined via the KL Divergence between the joint distribution of the variables and the product of their marginals: $I(X, Y) = D_{KL}(P(XY)|| P(X)P(Y))$.
For notational simplicity, we describe the joint distribution as $P$ and the product of the marginals as $Q$.
Let us further state that $P$ and $Q$ define outputs that are jointly in $R^D$.

A lower-bound for the KL divergence is as follows, setting $F$ as any class of functions that maps from $R^D$ to $R$:

\begin{equation}
    D_{KL}(P|| Q) \geq \mbox{sup}_{T \in F} \EX_P [T] - \mbox(\EX_Q[e^T])
    \label{eq:mine}
\end{equation}

In other words, one can lower bound the mutual information by finding a function, $T$, that maximizes the difference between the two terms in Equation~\ref{eq:mine}.
\citet{mine} do so with functions parametrized as a neural net that maps from the concatenation of two inputs (one for each random variable) to a single-valued output.
Training the neural net is conducted to maximize the value described by Equation~\ref{eq:mine}.

In our experiments, we create neural networks with separate, linear layers of size 64 for each input.
The embeddings from those two layers are concatenated, passed through two 1024-dimensional layers with ReLU activations, and then passed through a linear layer with a single output.
We thus mapedp from the two inputs to a single, real-valued output.

Training was performed using batch size 32 over 50 epochs, at which point the mutual information estimates appeared to have converged.

\section{Test Suite Creation}
\label{app:suites}
Here, we specify the details of the test suites used to evaluate models for reproducibility.

The Mask model Coordination test suite comprised sentences like ``The man saw the girl and the dog [MASK] tall.''
More generally, sentences followed the following template: ``The NN1 V the NN2 and the NN3 [MASK] ADJ.''
We created all sentences by iterating through the combinations of the words described in Table~\ref{tab:template_mlm_coord}.
This generated 243 sentences, and each sentence was associated with 2 parses: one described as a conjunction of sentences (e.g., ``(The man saw the girl) and (the dog [MASK] tall.)'') and one as a single sentence with a conjunction of noun phrases (e.g., ``The man saw (the girl and the dog) [MASK] tall.'').

\begin{table}[htb]
    \centering
    \begin{tabular}{c|l}
        Category & Words\\
        \hline
         NN1 & man, woman, child \\
         NN2 & boy, building, cat \\
         NN3 & dog, girl, truck \\
         V & saw, feared, heard \\
         ADJ & tall, falling, orange
    \end{tabular}
     \caption{Words used for sentence generation in the Mask Coordination test suite.}
    \label{tab:template_mlm_coord}
\end{table}

The mask model NP/Z test suite comprised sentences like, ``When the dog scratched the vet [MASK] ran.''
More generally, sentences followed the following template: ``When the NN1 V1 the NN2 [MASK] V2.''
Each sentence was associated with two parses, favoring either adverbs (e.g., ''When the dog scratched the vet quickly ran'' or nouns, ``When the dog scratch the vet she ran'').
We used the word tuples described in Table~\ref{tab:template_mlm_npz}, inspired by prior art, to generate 150 sentences.

\begin{table*}[htb]
    \centering
    \begin{tabular}{llll}
         NN1 & V1 & NN2 & V2\\
        \hline
         (dog/child) & (scratched/bit) & (vet/girl/boy) & (ran/screamed/smiled) \\
         author & wrote & book & grew \\
         (doctor/professor) & lectured & student & listened \\
         (girls/boys) & raced & (kids/children) & (watched/cheered) \\
         (people/spectators) & watched & (show/movie) & (stopped/paused)\\
         (lawyers/judges) & (studied/considered) & case & (languished/proceeded)\\
         (people/viewers) & (notice/spot) & actor & (departs/stays)\\
         (band/conventions) & left & (hotel/stalls) & closed
    \end{tabular}
     \caption{Words used for sentence generation in the Mask NP/Z test suite.}
    \label{tab:template_mlm_npz}
\end{table*}

The QA model Coordination test suite comprised prompts like ``Who was tall? The happy stranger saw the angry men and the angry women were tall.''
More generally, the prompts followed the following template: ``Who was ADJ1? The ADJ2 NN1 V the ADJ3 NN3 and the ADJ4 NN4 were ADJ1.''
We created 256 prompts by iterating through combinations of the words in Table~\ref{tab:template_qa_coord}.
``None'' adjectives were excluded from the text.

\begin{table}[htb]
    \centering
    \begin{tabular}{c|l}
        Category & Words\\
        \hline
         ADJ1 & tall, short \\
         ADJ2 & happy, None \\
         ADJ3 & angry, None \\
         ADJ4 & angry, None \\
         NN1 & stranger, child \\
         NN2 & men, women \\
         NN3 & women, men \\
         V & saw, believed
    \end{tabular}
     \caption{Words used for sentence generation in the QA Coordination test suite.}
    \label{tab:template_qa_coord}
\end{table}

The QA model NP/VP suite comprised prompts like ``Who had the telescope? The girl saw the boy with the telescope.''
The prompts followed the following template: ``Who had the NN1? The ADJ1 NN2 ADV V the ADJ2 NN3 with the ADJ3 NN4.''
In this suite, the choice of V and NN4 was tightly coupled - one may see with a telescope but not see with a stick, for example.
Table~\ref{tab:template_qa_npvp} details the combinations of words used to fill out the template, including V-NN4 pairs.
Overall, we generated 256 prompts.

\begin{table}[htb]
    \centering
    \begin{tabular}{c|l}
        Category & Words\\
        \hline
        V - NN4 & (saw, telescope), (poked, stick)\\
         ADJ1 & tall, None \\
         ADJ2 & short, None \\
         ADJ3 & special, None \\
         NN1 & man, woman \\
         NN2 & boy, girl \\
    \end{tabular}
     \caption{Words used for sentence generation in the QA NP/VP test suite.}
    \label{tab:template_qa_npvp}
\end{table}

The QA model RC suite comprised prompts like ``Who was desperate? The women and the men who were desperate bribed the politician.''
The prompts followed the following template: ``Who was ADJ1? The ADJ2 NN1 and the ADJ3 NN2 who were ADJ1 V the NN3.''
We generated 192 example prompts by iterating over combinations of the words listed in Table~\ref{tab:template_qa_rc}, excluding sentences in which NN1 and NN2 or ADJ2 and ADJ3 would have been the same.

\begin{table}[htb]
    \centering
    \begin{tabular}{c|l}
        Category & Words\\
        \hline
         ADJ1 & corrupt, desperate \\
         ADJ2 & tall, smart, rich \\
         ADJ3 & tall, smart, rich \\
         NN1 & men, women \\
         NN2 & men, women \\
         NN3 & judge, politician \\
    \end{tabular}
     \caption{Words used for sentence generation in the QA RC test suite.}
    \label{tab:template_qa_rc}
\end{table}

The Intervention suite for the QA model comprised prompts like ``What was green? The human saw the keys by the cabinet which were green.''
More generally, prompts were created via the following template: ``What was ADJ1? The NN1 V the NN2 by the NN3 which was/were ADJ1.''
By changing the plurality of NN2 or NN3 and replacing ``was'' with ``were,'' the correct answer should change.
Overall, we generated 288 sentences by iterating over all combinations of the words listed in Table~\ref{tab:template_qa_inter}, such that exactly one of NN1 and NN2 was plural at a time.

\begin{table}[htb]
    \centering
    \begin{tabular}{c|l}
        Category & Words\\
        \hline
         ADJ1 & green, large, dirty \\
         NN1 & human, stranger, child \\
         NN2 & key, keys, gadget, gadgets \\
         NN3 & cabinet, cabinets, vase, vases \\
    \end{tabular}
     \caption{Words used for sentence generation in the QA intervention experiments.}
    \label{tab:template_qa_inter}
\end{table}

\section{Hyperparameter Selection}
\label{app:hypers}

In the intervention experiments in Section~\ref{sec:boosting}, we performed interventions at layer 4, based on results of a validation study shown below.
We reported the results for probes with different dropout rates and for varying counterfactual losses, but we had to choose the layer of the QA model at which to perform interventions.

Therefore, we created a validation suite based on the Intervention template, using new nouns, verbs, and adjectives.
For dropout rates from 0.0 to 0.3, ranging over counterfactual losses, and layers from 1 to 7, we computed the QA model's F1 and Exact Match scores on the validation suite.
These results are included in Table~\ref{tab:qa_intervene}, and strongly suggested that performance, for all probes, was most increased via interventions at layer 4.

\begin{table*}[htb]
    \centering
    \begin{tabular}{c|ccccc}
         $\alpha$/Loss & Layer & 0.05 & 0.1 & 0.2 & 0.3\\
         \hline
         \multirow{7}{*}{Dist. 0.0}
         &1 & 71.9/59.4 & 72.7/60.9 & 73.4/60.9 & 73.4/60.9 \\
         &2 & 69.5/56.3 & 71.9/60.9 & 71.9/60.9 & 71.9/59.4\\
         &3 & 71.1/60.9 & 71.1/59.4 & 71.9/59.4 & 71.9/59.4\\
         &\textbf{4} & \textbf{71.9/62.5} & \textbf{72.6/60.4} & \textbf{71.9/59.4} & \textbf{73.4/60.9}\\
         &5 & 68.8/57.8 & 68.8/56.3 & 72.7/60.9 & 73.4/62.5\\
         &6 & 68.8/57.8 & 69.5/59.4 & 71.9/60.9 & 72.7/62.5\\
         &7 & 70.3/60.9 & 70.3/60.9 & 72.6/62.5 & 72.6/62.5\\
         \hline
         \multirow{7}{*}{Dist. 0.1}
         &1 & 69.5/56.3 & 71.1/59.4 & 71.9/59.4 & 71.9/59.4\\
         &2 & 68.8/60.9 & 70.3/60.9 & 69.3/59.4 & 71.1/59.4\\
         &3 & 67.2/56.4 & 69.5/60.9 & 72.7/60.9 & 73.4/62.5\\
         &\textbf{4} & \textbf{75.8/64.1} & \textbf{72.7/60.9} & \textbf{72.7/60.9} & \textbf{72.7/60.9}\\
         &5 & 68.8/56.3 & 70.3/59.4 & 71.9/57.8 & 71.1/56.3\\
         &6 & 75.0/59.4 & 72.7/60.9 & 73.4/62.5 & 73.4/62.5\\
         &7 & 72.7/60.9 & 72.7/62.5 & 72.7/62.5 & 72.7/60.9\\
         \hline
         \multirow{7}{*}{Dist. 0.2}
         &1 & 69.5/54.7 & 70.3/56.3 & 72.7/59.4 & 73.4/60.9\\
         &2 & 73.4/60.9 & 74.2/59.4 & 74.2/62.5 & 74.2/62.5\\
         &3 & 70.3/59.4 & 69.5/56.3 & 71.1/57.8 & 71.9/57.8\\
         &\textbf{4} & \textbf{74.2/65.6} & \textbf{75.0/65.6} & \textbf{75.8/65.6} & \textbf{75.0/64.1}\\
         &5 & 71.1/62.5 & 71.9/64.1 & 71.1/62.5 & 71.9/60.9\\
         &6 & 73.4/62.5 & 71.8/59.4 & 74.2/62.5 & 74.2/62.5\\
         &7 & 71.9/59.4 & 73.4/62.5 & 72.7/62.5 & 72.7/60.9\\
         \hline
         \multirow{7}{*}{Dist. 0.3}
         &1 & 67.2/54.7 & 70.3/59.4 & 73.4/62.5 & 72.7/60.9\\
         &2 & 68.8/60.9 & 71.1/60.9 & 72.7/62.5 & 71.9/60.9\\
         &3 & 61.7/53.1 & 64.8/56.3 & 71.9/64.1 & 72.3/65.6\\
         &\textbf{4} & \textbf{67.2/59.4} & \textbf{71.9/64.1} & \textbf{75.0/65.6} & \textbf{75.8/65.6}\\
         &5 & 62.5/56.3 & 68.8/59.4 & 70.3/62.5 & 70.3/62.5\\
         &6 & 71.1/62.5 & 711/64.1 & 70.3/60.9 & 71.1/60.9\\
         &7 & 75.0/64.1 & 72.7/62.5 & 71.9/62.5 & 73.4/62.5
    \end{tabular}
    \caption{Validation Coord. suite results (F1/Exact Match) using distance probes. For each probe type, we iterated over intervention layer and counterfactual loss value. The small validation suite was useful for rapid identification of good hyperparameter settings. All probes had the best performance at layer 4 (in bold).}
    \label{tab:qa_intervene}
\end{table*}

\section{Varying Dropout Rates}
\label{app:varying}
In the main paper, we reported included only some of the results for distance- and depth-based probe interventions.
Here, we first show, in more detail, how increasing the dropout rate grows the causal effect with the QA attachment suite and distance probes of varying $\alpha$.
Next, we include the mean causal effect plots for Mask and QA models using both types of probes on the 5 total suites.

\begin{figure*}[!htb]
    \setlength\tabcolsep{1pt}
    \settowidth\rotheadsize{Distance}
    \setkeys{Gin}{width=\hsize}
    \centering
    QA Model Causal Effect on Attachment Suites Using Dropout Distance Probes
    \begin{tabularx}{\linewidth}{sX}
        & \\
        \rothead{\centering Dropout 0.0} &
            \includegraphics[trim={0cm 1.5cm 0.2cm 0.0cm}, clip=true, valign=m]{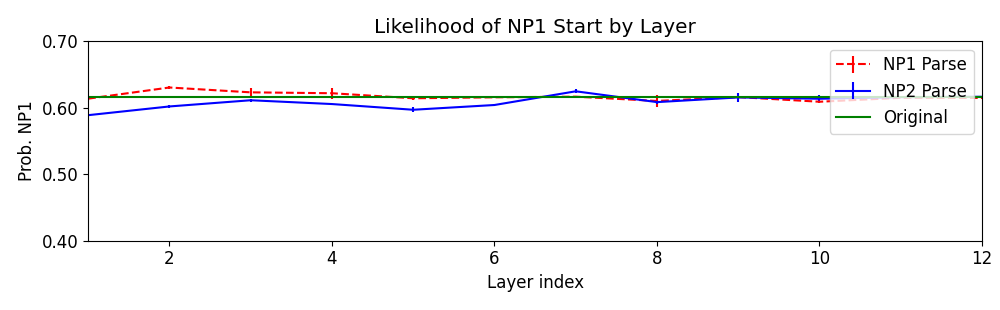} \\  
        \addlinespace[1pt]
        \rothead{\centering Dropout 0.1} &
            \includegraphics[trim={0cm 1.5cm 0.2cm 0.85cm}, clip=true, valign=m]{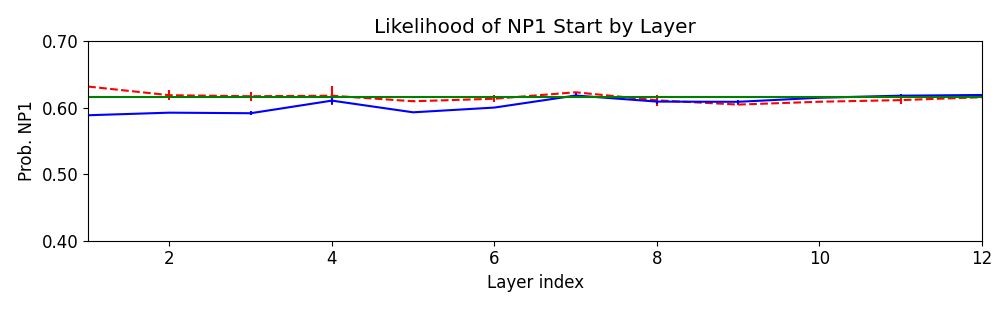} \\ 
        \addlinespace[1pt]
        \rothead{\centering Dropout 0.2} &
            \includegraphics[trim={0cm 1.5cm 0.2cm 0.85cm}, clip=true, valign=m]{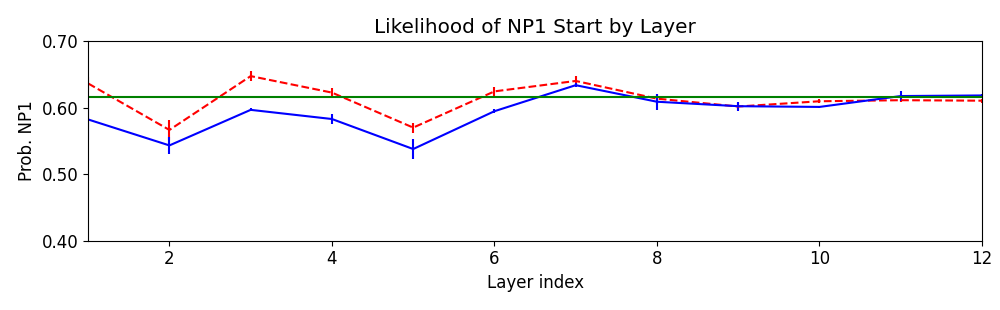} \\ 
        \addlinespace[1pt]
        \rothead{\centering Dropout 0.3} &
            \includegraphics[trim={0cm 1.5cm 0.2cm 0.85cm}, clip=true, valign=m]{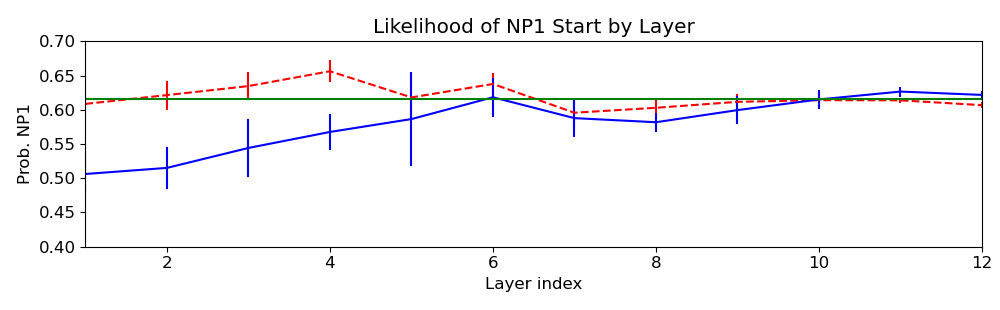} \\ 
        \addlinespace[1pt]
        \rothead{\centering Dropout 0.4} &
            \includegraphics[trim={0cm 1.4cm 0.2cm 0.85cm}, clip=true, valign=m]{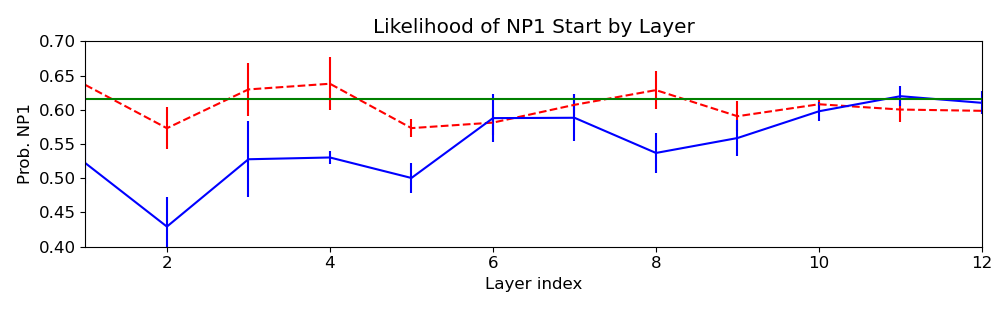} \\ 
        \addlinespace[1pt]
        \rothead{\centering Dropout 0.5} &
            \includegraphics[trim={0cm 1.4cm 0.2cm 0.85cm}, clip=true, valign=m]{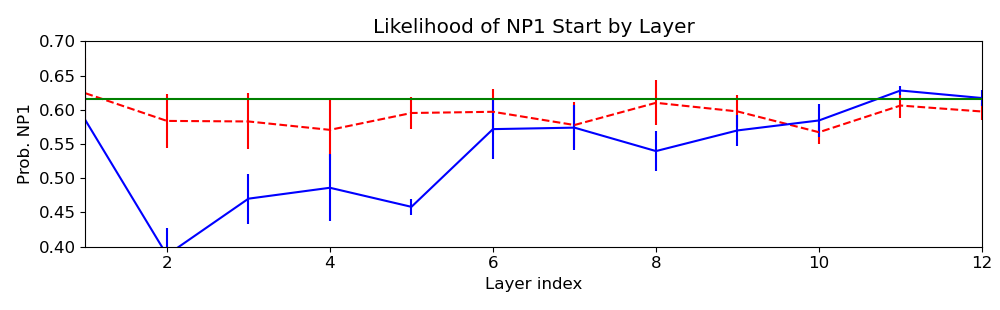} \\ 
    \end{tabularx}
    \caption{Dropout distance probes with dropout rates from 0.0 to 0.5 showed how, to a point, increasing the dropout rate increased the effect size for QA models on the Coord. suite.}
    \label{fig:varying}
\end{figure*}

First, we plotted an example of how increasing the dropout rate grew the causal effect in the QA attachment quite in Figure~\ref{fig:varying}.
We found that positive dropout values consistently outperformed probes with no dropout.
Furthermore, for $\alpha$ ranging from 0.1 to 0.4, increasing the dropout rate seemed to increase the effect size.
Considering only interventions at layer 2, for example, vanilla probes shifted model predictions by at most 2\% for different parses; for probes with dropout 0.5, probabilities shifted by roughly 20\%.

\begin{figure*}[!htb]
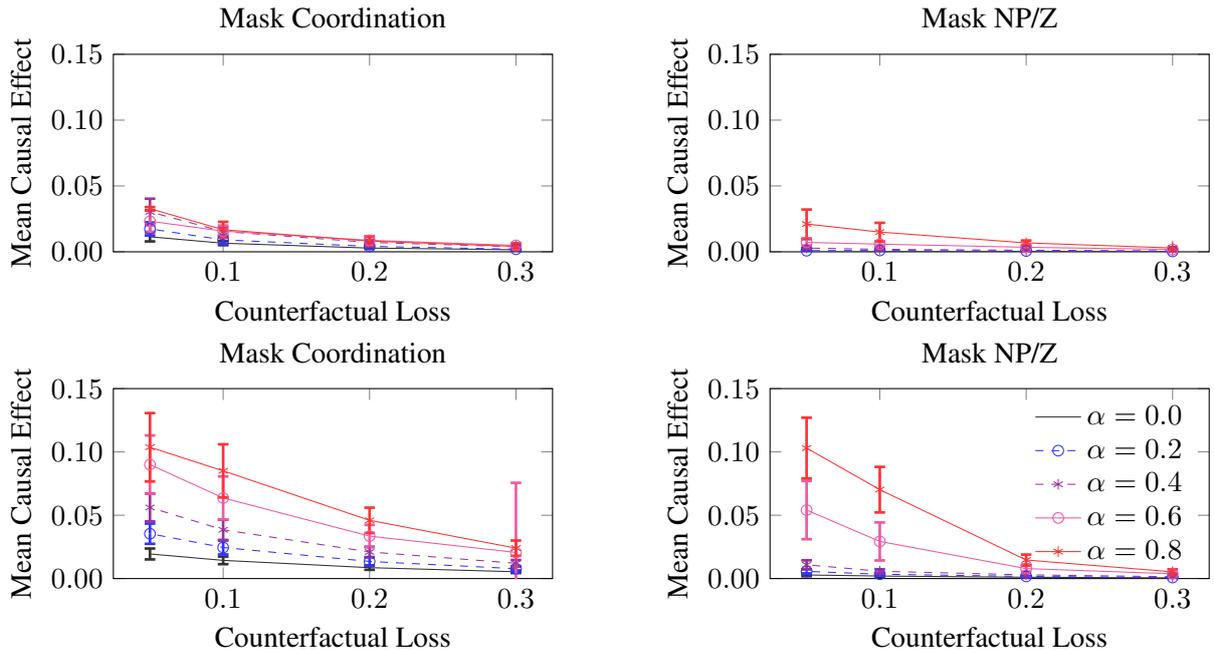

    \centering
    \begin{subfigure}[b]{0.46\textwidth}
        \centering
        \includestandalone[trim={0.12cm 0.0cm 0.0cm 0.0cm}, clip=true]{figures/tikz_diagrams/mask_coord_depth}
    \end{subfigure}
    \hfill
    \begin{subfigure}[b]{0.46\textwidth}
        \centering
        \includestandalone[trim={0.12cm 0.0cm 0.0cm 0.0cm}, clip=true]{figures/tikz_diagrams/mask_npz_depth}
    \end{subfigure}
    \begin{subfigure}[b]{0.46\textwidth}
        \centering
        \includestandalone[trim={0.12cm 0.0cm 0.0cm 0.0cm}, clip=true]{figures/tikz_diagrams/mask_coord_dist}
    \end{subfigure}
    \hfill
    \begin{subfigure}[b]{0.46\textwidth}
        \centering
        \includestandalone[trim={0.12cm 0.0cm 0.0cm 0.0cm}, clip=true]{figures/tikz_diagrams/mask_npz_dist}
    \end{subfigure}
    \caption{Mask mean causal effects using depth- (top) or distance-based (bottom) probes. Depth probes revealed smaller effects than distance-based probes, but a similar pattern of benefiting from lower counterfactual loss and higher dropout.}
    \label{fig:mask_plots_app}
\end{figure*}

\begin{figure*}[!htb]
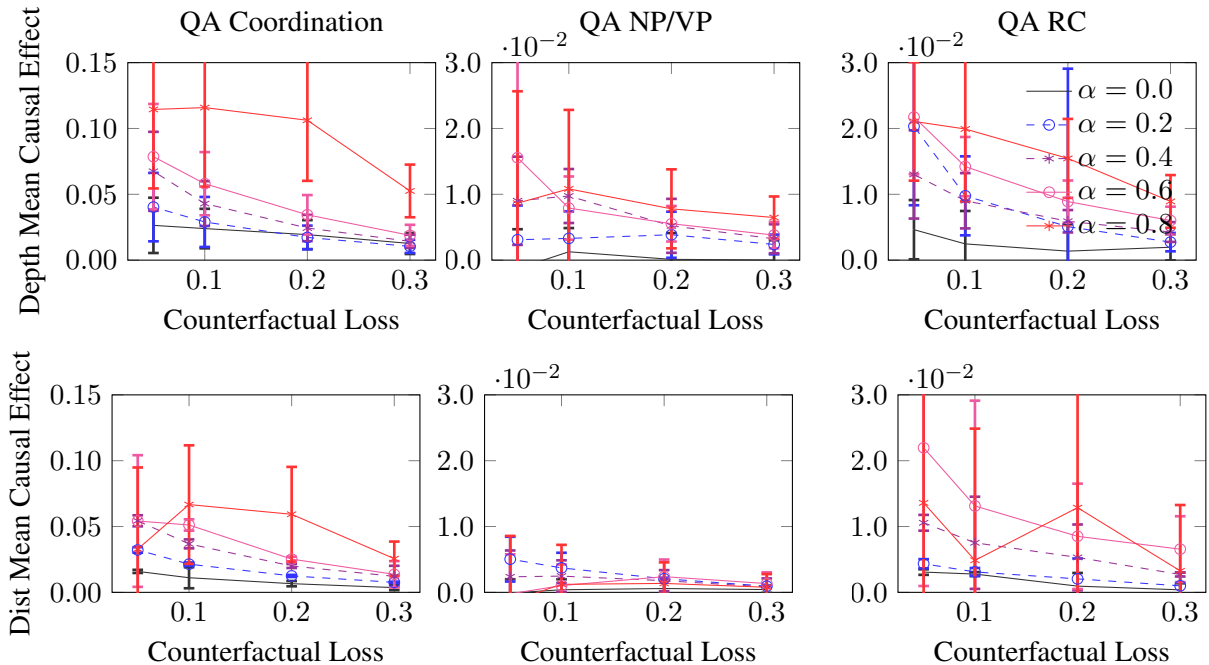

    \centering
    
    \begin{subfigure}[b]{0.32\textwidth}
        \includestandalone[trim={0.12cm 0.0cm 0.0cm 0.0cm}, clip=true]{figures/tikz_diagrams/qa_coord_depth_copied}
    \end{subfigure}
    ~
    \begin{subfigure}[b]{0.32\textwidth}
        \centering
        \includestandalone[trim={0.25cm 0.0cm 0.0cm 0.0cm}, clip=true]{figures/tikz_diagrams/qa_npvp_depth}
    \end{subfigure}
    \begin{subfigure}[b]{0.32\textwidth}
        \centering
        \includestandalone[trim={0.25cm 0.0cm 0.0cm 0.0cm}, clip=true]{figures/tikz_diagrams/qa_rc_depth}
    \end{subfigure}
    
    \begin{subfigure}[b]{0.32\textwidth}
        \centering
        \includestandalone[trim={0.12cm 0.0cm 0.0cm 0.0cm}, clip=true]{figures/tikz_diagrams/qa_coord_dist}
    \end{subfigure}
    \hfill
    \begin{subfigure}[b]{0.32\textwidth}
        \centering
        \includestandalone[trim={0.25cm 0.0cm 0.0cm 0.0cm}, clip=true]{figures/tikz_diagrams/qa_npvp_dist}
    \end{subfigure}
    \hfill
    \begin{subfigure}[b]{0.32\textwidth}
        \centering
        \includestandalone[trim={0.25cm 0.0cm 0.0cm 0.0cm}, clip=true]{figures/tikz_diagrams/qa_rc_dist}
    \end{subfigure}
    \caption{QA mean causal effects using depth-based (top) or distance-based (bottom) probes.}
    \label{fig:qa_plots_app}
\end{figure*}

Finally, we included results for all dropout rates and counterfactual losses in Figures~\ref{fig:mask_plots_app} and \ref{fig:qa_plots_app}.

\section{Probe Performance Metrics}
\label{app:probe_perf}
In the main paper, we demonstrated the benefits of using dropout probes for creating counterfactual embeddings.
One could hypothesize that the dropout enables better counterfactuals because the probes are prevented from overfitting to the training data.
We found that that was not the case.

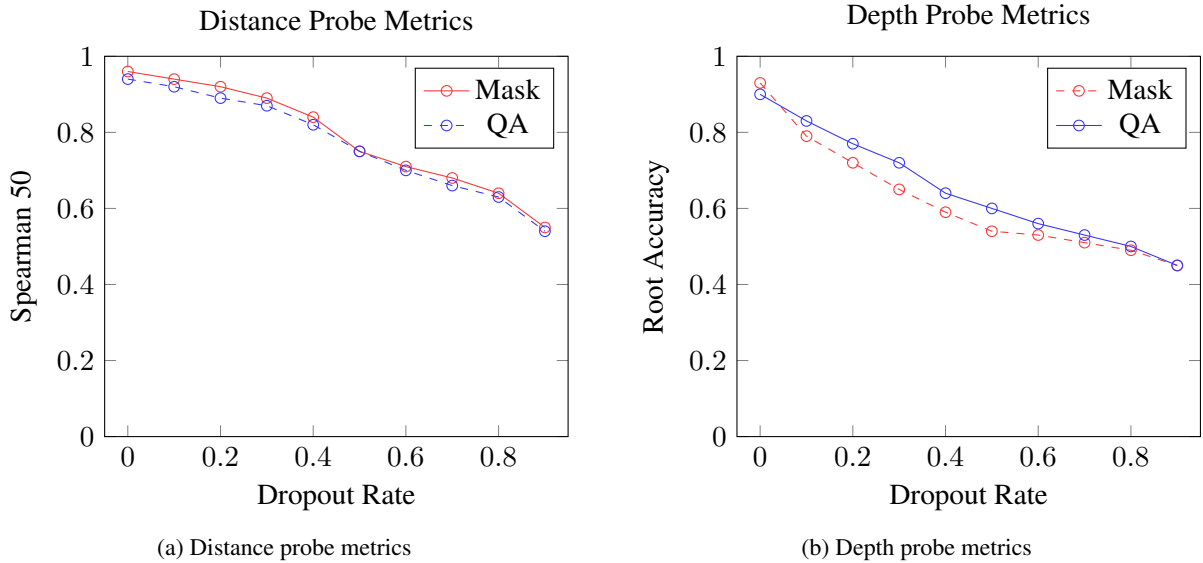
\begin{figure*}[ht]
    \centering
    \begin{subfigure}[b]{0.48\textwidth}
        \centering
        \begin{tikzpicture}[scale=1.0] 
            \begin{axis}[legend pos=north east,  width={\textwidth},
                ylabel=Spearman 50,
    xlabel=Dropout Rate,
    xmin=-0.05, xmax=0.95,
    ymin=0.0,ymax=1.0,
    title={Distance Probe Metrics}]
            \addplot [mark=o, red!80, 
                      mark options={solid}]
                    table [x=x, y=y, y error=y-err]{
                        x	y y-err
                        0.0	0.96 0.00
                        0.1	0.94 0.00
                        0.2	0.92 0.00
                        0.3	0.89 0.00
                        0.4	0.84 0.01
                        0.5	0.75 0.00
                        0.6	0.71 0.01
                        0.7	0.68 0.00
                        0.8	0.64 0.00
                        0.9	0.55 0.00
                            };
            \addlegendentry{Mask}
            \addplot [mark=o, blue!80, dashed, 
                      mark options={solid}]
                    table [x=x, y=y, y error=y-err]{
                        x	y y-err
                        0.0	0.94 0.00
                        0.1	0.92 0.00
                        0.2	0.89 0.00
                        0.3	0.87 0.00
                        0.4	0.82 0.01
                        0.5	0.75 0.01
                        0.6	0.70 0.01
                        0.7	0.66 0.00
                        0.8	0.63 0.00
                        0.9	0.54 0.00
                            };
            \addlegendentry{QA}

            \end{axis} 
        \end{tikzpicture}
        \caption{Distance probe metrics}
    \end{subfigure}
    \hfill
    \begin{subfigure}[b]{0.48\textwidth}
        \centering
        \begin{tikzpicture}[scale=1.0] 
            \begin{axis}[legend pos=north east,  width={\textwidth},
                ylabel=Root Accuracy,
    xlabel=Dropout Rate,
    xmin=-0.05, xmax=0.95,
    ymin=0.0,ymax=1.0,
    title={Depth Probe Metrics}]
            \addplot [mark=o, red!80, dashed,
                      mark options={solid}]
                    table [x=x, y=y, y error=y-err]{
                        x	y y-err
                        0.0	0.93 0.00
                        0.1	0.79 0.00
                        0.2	0.72 0.01
                        0.3	0.65 0.01
                        0.4	0.59 0.01
                        0.5	0.54 0.01
                        0.6	0.53 0.01
                        0.7	0.51 0.01
                        0.8	0.49 0.00
                        0.9	0.45 0.00
                            };
            \addlegendentry{Mask}
            \addplot [mark=o, blue!80, 
                      mark options={solid}]
                    table [x=x, y=y, y error=y-err]{
                        x	y y-err
                        0.0	0.90 0.00
                        0.1	0.83 0.00
                        0.2	0.77 0.00
                        0.3	0.72 0.00
                        0.4	0.64 0.00
                        0.5	0.60 0.01
                        0.6	0.56 0.01
                        0.7	0.53 0.01
                        0.8	0.50 0.00
                        0.9	0.45 0.00
                            };
            \addlegendentry{QA}

            \end{axis} 
        \end{tikzpicture}
        \caption{Depth probe metrics}
    \end{subfigure}
    \caption{Metrics for the distance (left) and depth (right) probes showed that introducing dropout worsened probe performance as measured on the probe prediction tasks. Means over 5 trials plotted. All standard deviations less than 0.01.}
    \label{fig:dist_probe_perf}
\end{figure*}

In Figure~\ref{fig:dist_probe_perf}, we plotted probe performance metrics for the distance- and depth-based probes.
For the distance probe, we reported the spearman correlation coefficient between predicted and actual pairwise distances between words in a sentence's parse tree.
For the depth probe, we reported the accuracy of the probe in predicting the word at the root of the syntax tree.
Both metrics were used in prior probing literature \cite{hewitt2019structural}.

We found that, while using non-linear probes boosted probe performance compared to linear probes, adding dropout actually worsened probe performance.
This suggests that the benefits from dropout in counterfactual generation arose from a phenomenon other than higher-performing probes.

\section{Scientific Artifacts}
In this work, we built upon pre-existing scientific artifacts, including datasets and publicly-avaible code.
Here, we briefly list their licenses and intended use cases.
We used all artifacts for purely academic purposes.

The Penn TreeBank is licensed under the ``LDC User Agreement for Non-Members'' \cite{ptb}.
The dataset is commonly used in many academic settings (e.g., \citet{hewitt2019structural,whatif}).

The Stanford Question Answering Dataset 2.0 is under a creative commons license and is commonly used in academic settings \cite{squad}.

The MNLI dataset is under an OANC license, ``which allows all content to be freely used, modified, and shared under permissive terms'' \cite{mnli}.
The HANS dataset is under an MIT license \cite{rightwrong}.
Both datasets are commonly used in academic settings \cite{rightwrong}.

The code we used to train the NLI and NLI-HANS models is under an Apache License 2.0 \cite{bert-nli-code} and was developed for academic use.

\end{document}

%% file: figures/tikz_diagrams/mask_coord_dist.tex
        \begin{tikzpicture}[scale=1.0] 
            \begin{axis}[legend pos=north east,  width={\textwidth},
                ylabel=Mean Causal Effect,
    xlabel=Counterfactual Loss,
    xmin=0.025, xmax=0.325,
    ymin=0.0,ymax=0.15,
    title={Mask Coordination},
    height=4.1cm,
    y tick label style={
        /pgf/number format/.cd,
        fixed,
        fixed zerofill,
        precision=2,
        /tikz/.cd
    }]
            \addplot [black!80, 
                      mark options={solid},
                      error bars/.cd, y dir=both, y explicit,
                      error bar style={line width=1pt,solid},
                      error mark options={line width=1pt,mark size=2pt,rotate=90}]
                    table [x=x, y=y, y error=y-err]{
                        x	y y-err
                        0.3	    0.005355776348 0.001
                        0.2	    0.008609295832 0.0017
                        0.1	    0.0143531141 0.003
                        0.05    0.01941898604 0.0043
                            };
            \addplot [mark=o, blue!80, dashed, 
                      mark options={solid},
                      error bars/.cd, y dir=both, y explicit,
                      error bar style={line width=1pt,solid},
                      error mark options={line width=1pt,mark size=2pt,rotate=90}]
                    table [x=x, y=y, y error=y-err]{
                        x	y y-err
                        0.3 	0.007779071681 0.002
                        0.2 	0.01351172002 0.0033
                        0.1 	0.02453508592 0.0055
                        0.05	0.0354193886 0.008
                            };
            \addplot [mark=asterisk, violet!80, dashed, 
                      mark options={solid},
                      error bars/.cd, y dir=both, y explicit,
                      error bar style={line width=1pt,solid},
                      error mark options={line width=1pt,mark size=2pt,rotate=90}]
                    table [x=x, y=y, y error=y-err]{
                        x	y y-err
                        0.3 	0.01205728004 0.0026
                        0.2 	0.02102689986 0.0041
                        0.1 	0.03851081876 0.008
                        0.05	0.05610721433 0.011
                            };
            \addplot [mark=o, magenta!80,
                      mark options={solid},
                      error bars/.cd, y dir=both, y explicit,
                      error bar style={line width=1pt,solid},
                      error mark options={line width=1pt,mark size=2pt,rotate=90}]
                    table [x=x, y=y, y error=y-err]{
                        x	y y-err
                        0.3 	0.02062124311 0.055
                        0.2 	0.03348774247 0.0089
                        0.1 	0.0636545126 0.017
                        0.05	0.09000149494 0.023
                            };
            \addplot [mark=asterisk, red!80,
                      mark options={solid},
                      error bars/.cd, y dir=both, y explicit,
                      error bar style={line width=1pt,solid},
                      error mark options={line width=1pt,mark size=2pt,rotate=90}]
                    table [x=x, y=y, y error=y-err]{
                        x	y y-err
                        0.3 	0.02400744036 0.006
                        0.2 	0.04597768435 0.01
                        0.1 	0.08505637582 0.021
                        0.05	0.1036943692 0.027
                            };
            \end{axis} 
        \end{tikzpicture}

%% file: figures/tikz_diagrams/mask_npz_dist.tex
        \begin{tikzpicture}[scale=1.0] 
            \begin{axis}[legend pos=north east,  width={\textwidth},
                ylabel=Mean Causal Effect,
      legend style={fill=none, draw=none},
    xlabel=Counterfactual Loss,
    xmin=0.025, xmax=0.325,
    ymin=0.0,ymax=0.15,
    title={Mask NP/Z},
    height=4.1cm,
    y tick label style={
        /pgf/number format/.cd,
        fixed,
        fixed zerofill,
        precision=2,
        /tikz/.cd
    }]
            \addplot [black!80, 
                      mark options={solid},
                      error bars/.cd, y dir=both, y explicit,
                      error bar style={line width=1pt,solid},
                      error mark options={line width=1pt,mark size=2pt,rotate=90}]
                    table [x=x, y=y, y error=y-err]{
                        x	y y-err
                        0.3	    0.0005453077785 0.0002
                        0.2	    0.00104487217 0.00035
                        0.1	    0.001978419054 0.0007
                        0.05    0.002836904186 0.001
                            };
            \addlegendentry{$\alpha = 0.0$}
            \addplot [mark=o, blue!80, dashed, 
                      mark options={solid},
                      error bars/.cd, y dir=both, y explicit,
                      error bar style={line width=1pt,solid},
                      error mark options={line width=1pt,mark size=2pt,rotate=90}]
                    table [x=x, y=y, y error=y-err]{
                        x	y y-err
                        0.3 	0.0008247844836 0.0002
                        0.2 	0.00168163396 0.0005
                        0.1 	0.00341501486 0.001
                        0.05	0.00565714048 0.0017
                            };
            \addlegendentry{$\alpha = 0.2$}
            \addplot [mark=asterisk, violet!80, dashed, 
                      mark options={solid},
                      error bars/.cd, y dir=both, y explicit,
                      error bar style={line width=1pt,solid},
                      error mark options={line width=1pt,mark size=2pt,rotate=90}]
                    table [x=x, y=y, y error=y-err]{
                        x	y y-err
                        0.3 	0.001448084681 0.0004
                        0.2 	0.002921164832 0.008
                        0.1 	0.005768721821 0.0016
                        0.05	0.01087628169 0.0035
                            };
            \addlegendentry{$\alpha = 0.4$}
            \addplot [mark=o, magenta!80,
                      mark options={solid},
                      error bars/.cd, y dir=both, y explicit,
                      error bar style={line width=1pt,solid},
                      error mark options={line width=1pt,mark size=2pt,rotate=90}]
                    table [x=x, y=y, y error=y-err]{
                        x	y y-err
                        0.3 	0.003761252568 0.001
                        0.2 	0.007879548252 0.0024
                        0.1 	0.02930383599 0.015
                        0.05	0.05407501653 0.023
                            };
            \addlegendentry{$\alpha = 0.6$}
            \addplot [mark=asterisk, red!80,
                      mark options={solid},
                      error bars/.cd, y dir=both, y explicit,
                      error bar style={line width=1pt,solid},
                      error mark options={line width=1pt,mark size=2pt,rotate=90}]
                    table [x=x, y=y, y error=y-err]{
                        x	y y-err
                        0.3 	0.005316997071 0.002
                        0.2 	0.01463993204 0.0043
                        0.1 	0.0702198484 0.018
                        0.05	0.1030143393 0.024
                            };
            \addlegendentry{$\alpha = 0.8$}

            \end{axis} 
        \end{tikzpicture}

%% file: figures/tikz_diagrams/qa_coord_depth.tex
       \begin{tikzpicture}
            \begin{axis}[legend pos=north east,  width={1.1\textwidth},
                ylabel=Mean Causal Effect,
    xlabel=Counterfactual Loss,
    xmin=0.025, xmax=0.325,
    ymin=0.0,ymax=0.15,
    title={QA Coordination},
    height=4.2cm,
    y tick label style={
        /pgf/number format/.cd,
        fixed,
        fixed zerofill,
        precision=2,
        /tikz/.cd
    }]
            \addplot [black!80, 
                      mark options={solid},
                      error bars/.cd, y dir=both, y explicit,
                      error bar style={line width=1pt,solid},
                      error mark options={line width=1pt,mark size=2pt,rotate=90}]
                    table [x=x, y=y, y error=y-err]{
                        x	y y-err
                        0.3	    0.01265611355 0.008
                        0.2	    0.01937701014 0.011
                        0.1	    0.02397855954 0.015
                        0.05    0.02641605355 0.021
                            };
            \addplot [mark=o, blue!80, dashed, 
                      mark options={solid},
                      error bars/.cd, y dir=both, y explicit,
                      error bar style={line width=1pt,solid},
                      error mark options={line width=1pt,mark size=2pt,rotate=90}]
                    table [x=x, y=y, y error=y-err]{
                        x	y y-err
                        0.3 	0.01044824606 0.005
                        0.2 	0.0172309392 0.009
                        0.1 	0.02900734169 0.019
                        0.05	0.04028607925 0.026
                            };
            \addplot [mark=asterisk, violet!80, dashed, 
                      mark options={solid},
                      error bars/.cd, y dir=both, y explicit,
                      error bar style={line width=1pt,solid},
                      error mark options={line width=1pt,mark size=2pt,rotate=90}]
                    table [x=x, y=y, y error=y-err]{
                        x	y y-err
                        0.3 	0.01427423123 0.005
                        0.2 	0.02439524751 0.01
                        0.1 	0.04294733276 0.017
                        0.05	0.06745922205 0.030
                            };
            \addplot [mark=o, magenta!80,
                      mark options={solid},
                      error bars/.cd, y dir=both, y explicit,
                      error bar style={line width=1pt,solid},
                      error mark options={line width=1pt,mark size=2pt,rotate=90}]
                    table [x=x, y=y, y error=y-err]{
                        x	y y-err
                        0.3 	0.01883239322 0.008
                        0.2 	0.03442594473 0.015
                        0.1 	0.05805116958 0.024
                        0.05	0.07859456268 0.04
                            };
            \addplot [mark=asterisk, red!80,
                      mark options={solid},
                      error bars/.cd, y dir=both, y explicit,
                      error bar style={line width=1pt,solid},
                      error mark options={line width=1pt,mark size=2pt,rotate=90}]
                    table [x=x, y=y, y error=y-err]{
                        x	y y-err
                        0.3 	0.05256431142 0.02
                        0.2 	0.1062422903 0.046
                        0.1 	0.1158602629 0.06
                        0.05	0.1144569436 0.06
                            };
            \end{axis} 
        \end{tikzpicture}

%% file: figures/tikz_diagrams/qa_npvp_depth.tex
\setlength{\fboxsep}{0pt}
\begin{tikzpicture}
            \begin{axis}[legend pos=north east,  width={1.1\textwidth},
                xlabel=Counterfactual Loss,
                xmin=0.025, xmax=0.325,
                ymin=0.0,ymax=0.03,
                title={QA NP/VP},
                height=4.2cm,
                y tick label style={
                    /pgf/number format/.cd,
                    fixed,
                    fixed zerofill,
                    precision=1,
                    /tikz/.cd
                }]
            \addplot [black!80, 
                      mark options={solid},
                      error bars/.cd, y dir=both, y explicit,
                      error bar style={line width=1pt,solid},
                      error mark options={line width=1pt,mark size=2pt,rotate=90}]
                    table [x=x, y=y, y error=y-err]{
                        x	y y-err
                        0.3	    0.0000265 0.003
                        0.2	    0.000117 0.004
                        0.1	    0.001270581862 0.0036
                        0.05    -0.001304198387 0.006
                            };
            \addplot [mark=o, blue!80, dashed, 
                      mark options={solid},
                      error bars/.cd, y dir=both, y explicit,
                      error bar style={line width=1pt,solid},
                      error mark options={line width=1pt,mark size=2pt,rotate=90}]
                    table [x=x, y=y, y error=y-err]{
                        x	y y-err
                        0.3 	0.002381785296 0.0015
                        0.2 	0.003857980802 0.0035
                        0.1 	0.003301662321 0.0041
                        0.05	0.003098405509 0.0052
                            };
            \addplot [mark=asterisk, violet!80, dashed, 
                      mark options={solid},
                      error bars/.cd, y dir=both, y explicit,
                      error bar style={line width=1pt,solid},
                      error mark options={line width=1pt,mark size=2pt,rotate=90}]
                    table [x=x, y=y, y error=y-err]{
                        x	y y-err
                        0.3 	0.003246968361 0.0022
                        0.2 	0.005209454227 0.0041
                        0.1 	0.009740326548 0.0041
                        0.05	0.009013782971 0.0067
                            };
            \addplot [mark=o, magenta!80,
                      mark options={solid},
                      error bars/.cd, y dir=both, y explicit,
                      error bar style={line width=1pt,solid},
                      error mark options={line width=1pt,mark size=2pt,rotate=90}]
                    table [x=x, y=y, y error=y-err]{
                        x	y y-err
                        0.3 	0.003803104352 0.002
                        0.2 	0.005530027001 0.0027
                        0.1 	0.007910278166 0.0048
                        0.05	0.0155586551 0.016
                            };
            \addplot [mark=asterisk, red!80,
                      mark options={solid},
                      error bars/.cd, y dir=both, y explicit,
                      error bar style={line width=1pt,solid},
                      error mark options={line width=1pt,mark size=2pt,rotate=90}]
                    table [x=x, y=y, y error=y-err]{
                        x	y y-err
                        0.3 	0.006471410104 0.0032
                        0.2 	0.007791887106 0.006
                        0.1 	0.01082431568 0.012
                        0.05	0.008652800408 0.017
                            };
            \end{axis} 
        \end{tikzpicture}

%% file: figures/tikz_diagrams/qa_rc_depth.tex
\setlength{\fboxsep}{0pt}
\begin{tikzpicture}
            \begin{axis}[legend pos=north east,  width={1.1\textwidth},
    xlabel=Counterfactual Loss,
    xmin=0.025, xmax=0.325,
    ymin=0.0,ymax=0.03,
  legend style={fill=none, draw=none},
    title={QA RC},
    height=4.2cm,
    y tick label style={
        /pgf/number format/.cd,
        fixed,
        fixed zerofill,
        precision=1,
        /tikz/.cd
    }]
            \addplot [black!80, 
                      mark options={solid},
                      error bars/.cd, y dir=both, y explicit,
                      error bar style={line width=1pt,solid},
                      error mark options={line width=1pt,mark size=2pt,rotate=90}]
                    table [x=x, y=y, y error=y-err]{
                        x	y y-err
                        0.3	    0.00195896504044934 0.002
                        0.2	    0.00136845115389283 0.0041
                        0.1	    0.00246853145 0.005
                        0.05    0.00462828101 0.0045
                            };
            \addlegendentry{$\alpha = 0.0$}
            \addplot [mark=o, blue!80, dashed, 
                      mark options={solid},
                      error bars/.cd, y dir=both, y explicit,
                      error bar style={line width=1pt,solid},
                      error mark options={line width=1pt,mark size=2pt,rotate=90}]
                    table [x=x, y=y, y error=y-err]{
                        x	y y-err
                        0.3 	0.00273631844 0.0014
                        0.2 	0.005081653449 0.024
                        0.1 	0.009767199241 0.006
                        0.05	0.02033693833 0.012
                            };
            \addlegendentry{$\alpha = 0.2$}
            \addplot [mark=asterisk, violet!80, dashed, 
                      mark options={solid},
                      error bars/.cd, y dir=both, y explicit,
                      error bar style={line width=1pt,solid},
                      error mark options={line width=1pt,mark size=2pt,rotate=90}]
                    table [x=x, y=y, y error=y-err]{
                        x	y y-err
                        0.3 	0.004269707507 0.0015
                        0.2 	0.005906973067 0.0017
                        0.1 	0.009021856833 0.0042
                        0.05	0.01301507834 0.0067
                            };
            \addlegendentry{$\alpha = 0.4$}
            \addplot [mark=o, magenta!80,
                      mark options={solid},
                      error bars/.cd, y dir=both, y explicit,
                      error bar style={line width=1pt,solid},
                      error mark options={line width=1pt,mark size=2pt,rotate=90}]
                    table [x=x, y=y, y error=y-err]{
                        x	y y-err
                        0.3 	0.006018282478 0.0021
                        0.2 	0.008894344524 0.0032
                        0.1 	0.01420982435 0.0045
                        0.05	0.02173661235 0.0087
                            };
            \addlegendentry{$\alpha = 0.6$}
            \addplot [mark=asterisk, red!80,
                      mark options={solid},
                      error bars/.cd, y dir=both, y explicit,
                      error bar style={line width=1pt,solid},
                      error mark options={line width=1pt,mark size=2pt,rotate=90}]
                    table [x=x, y=y, y error=y-err]{
                        x	y y-err
                        0.3 	0.00892025719 0.004
                        0.2 	0.01545149128 0.006
                        0.1 	0.01992125268 0.011
                        0.05	0.02107257947 0.009
                            };
            \addlegendentry{$\alpha = 0.8$}
            \end{axis} 
        \end{tikzpicture}

%% file: figures/tikz_diagrams/mask_coord_depth.tex
        \begin{tikzpicture}[scale=1.0] 
            \begin{axis}[legend pos=north east,  width={\textwidth},
                ylabel=Mean Causal Effect,
    xlabel=Counterfactual Loss,
    xmin=0.025, xmax=0.325,
    ymin=0.0,ymax=0.15,
    title={Mask Coordination},
    height=4.2cm,
    y tick label style={
        /pgf/number format/.cd,
        fixed,
        fixed zerofill,
        precision=2,
        /tikz/.cd
    }]
            \addplot [black!80, 
                      mark options={solid},
                      error bars/.cd, y dir=both, y explicit,
                      error bar style={line width=1pt,solid},
                      error mark options={line width=1pt,mark size=2pt,rotate=90}]
                    table [x=x, y=y, y error=y-err]{
                        x	y y-err
                        0.3	    0.001558334212 0.0004
                        0.2	    0.002820127685 0.0006
                        0.1	    0.006517935538 0.0015
                        0.05    0.01139174793 0.0035
                            };
            \addplot [mark=o, blue!80, dashed, 
                      mark options={solid},
                      error bars/.cd, y dir=both, y explicit,
                      error bar style={line width=1pt,solid},
                      error mark options={line width=1pt,mark size=2pt,rotate=90}]
                    table [x=x, y=y, y error=y-err]{
                        x	y y-err
                        0.3 	0.001936925084 0.0005
                        0.2 	0.004115491562 0.0012
                        0.1 	0.009062597232 0.0032
                        0.05	0.01738531905 0.005
                            };
            \addplot [mark=asterisk, violet!80, dashed, 
                      mark options={solid},
                      error bars/.cd, y dir=both, y explicit,
                      error bar style={line width=1pt,solid},
                      error mark options={line width=1pt,mark size=2pt,rotate=90}]
                    table [x=x, y=y, y error=y-err]{
                        x	y y-err
                        0.3 	0.00379995288934908 0.0009
                        0.2 	0.007162775397 0.0015
                        0.1 	0.01539232708 0.0032
                        0.05	0.03030501668 0.01
                            };
            \addplot [mark=o, magenta!80,
                      mark options={solid},
                      error bars/.cd, y dir=both, y explicit,
                      error bar style={line width=1pt,solid},
                      error mark options={line width=1pt,mark size=2pt,rotate=90}]
                    table [x=x, y=y, y error=y-err]{
                        x	y y-err
                        0.3 	0.004844574719 0.0014
                        0.2 	0.008713379719 0.0025
                        0.1 	0.01571015876 0.0042
                        0.05	0.02318682863 0.008
                            };
            \addplot [mark=asterisk, red!80,
                      mark options={solid},
                      error bars/.cd, y dir=both, y explicit,
                      error bar style={line width=1pt,solid},
                      error mark options={line width=1pt,mark size=2pt,rotate=90}]
                    table [x=x, y=y, y error=y-err]{
                        x	y y-err
                        0.3 	0.004119538217 0.0023
                        0.2 	0.008117430331 0.0036
                        0.1 	0.01676496882 0.006
                        0.05	0.03274481544 0.0012
                            };
            \end{axis} 
        \end{tikzpicture}

%% file: figures/tikz_diagrams/mask_npz_depth.tex
        \begin{tikzpicture}[scale=1.0] 
            \begin{axis}[legend pos=north east,  width={\textwidth},
                ylabel=Mean Causal Effect,
    xlabel=Counterfactual Loss,
    xmin=0.025, xmax=0.325,
    ymin=0.0,ymax=0.15,
    title={Mask NP/Z},
    height=4.2cm,
    y tick label style={
        /pgf/number format/.cd,
        fixed,
        fixed zerofill,
        precision=2,
        /tikz/.cd
    }]
            \addplot [black!80, 
                      mark options={solid},
                      error bars/.cd, y dir=both, y explicit,
                      error bar style={line width=1pt,solid},
                      error mark options={line width=1pt,mark size=2pt,rotate=90}]
                    table [x=x, y=y, y error=y-err]{
                        x	y y-err
                        0.3	    0.0000878144529734836 0.0001
                        0.2	    0.0002138703515 0.00015
                        0.1	    -0.0000272666141206689 0.001
                        0.05    -0.0002120651565 0.0012
                            };
            \addplot [mark=o, blue!80, dashed, 
                      mark options={solid},
                      error bars/.cd, y dir=both, y explicit,
                      error bar style={line width=1pt,solid},
                      error mark options={line width=1pt,mark size=2pt,rotate=90}]
                    table [x=x, y=y, y error=y-err]{
                        x	y y-err
                        0.3 	0.0002894060399 0.00012
                        0.2 	0.0006047156801 0.00043
                        0.1 	0.00103730592 0.001
                        0.05	0.0008370565286 0.0012
                            };
            \addplot [mark=asterisk, violet!80, dashed, 
                      mark options={solid},
                      error bars/.cd, y dir=both, y explicit,
                      error bar style={line width=1pt,solid},
                      error mark options={line width=1pt,mark size=2pt,rotate=90}]
                    table [x=x, y=y, y error=y-err]{
                        x	y y-err
                        0.3 	0.0007431291419 0.0002
                        0.2 	0.001147456664 0.00036
                        0.1 	0.001900966495 0.0007
                        0.05	0.002822667469 0.00017
                            };
            \addplot [mark=o, magenta!80,
                      mark options={solid},
                      error bars/.cd, y dir=both, y explicit,
                      error bar style={line width=1pt,solid},
                      error mark options={line width=1pt,mark size=2pt,rotate=90}]
                    table [x=x, y=y, y error=y-err]{
                        x	y y-err
                        0.3 	0.001775046226 0.0003
                        0.2 	0.003408654414 0.0007
                        0.1 	0.005825226745 0.0018
                        0.05	0.007115635918 0.0023
                            };
            \addplot [mark=asterisk, red!80,
                      mark options={solid},
                      error bars/.cd, y dir=both, y explicit,
                      error bar style={line width=1pt,solid},
                      error mark options={line width=1pt,mark size=2pt,rotate=90}]
                    table [x=x, y=y, y error=y-err]{
                        x	y y-err
                        0.3 	0.002913457114 0.00056
                        0.2 	0.006761499684 0.0018
                        0.1 	0.01498855972 0.007
                        0.05	0.0210912717 0.011
                            };

            \end{axis} 
        \end{tikzpicture}

%% file: figures/tikz_diagrams/qa_coord_depth_copied.tex
       \begin{tikzpicture}
            \begin{axis}[legend pos=north east,  width={1.1\textwidth},
                ylabel=Depth Mean Causal Effect,
    xlabel=Counterfactual Loss,
    xmin=0.025, xmax=0.325,
    ymin=0.0,ymax=0.15,
    title={QA Coordination},
    height=4.2cm,
    y tick label style={
        /pgf/number format/.cd,
        fixed,
        fixed zerofill,
        precision=2,
        /tikz/.cd
    }]
            \addplot [black!80, 
                      mark options={solid},
                      error bars/.cd, y dir=both, y explicit,
                      error bar style={line width=1pt,solid},
                      error mark options={line width=1pt,mark size=2pt,rotate=90}]
                    table [x=x, y=y, y error=y-err]{
                        x	y y-err
                        0.3	    0.01265611355 0.008
                        0.2	    0.01937701014 0.011
                        0.1	    0.02397855954 0.015
                        0.05    0.02641605355 0.021
                            };
            \addplot [mark=o, blue!80, dashed, 
                      mark options={solid},
                      error bars/.cd, y dir=both, y explicit,
                      error bar style={line width=1pt,solid},
                      error mark options={line width=1pt,mark size=2pt,rotate=90}]
                    table [x=x, y=y, y error=y-err]{
                        x	y y-err
                        0.3 	0.01044824606 0.005
                        0.2 	0.0172309392 0.009
                        0.1 	0.02900734169 0.019
                        0.05	0.04028607925 0.026
                            };
            \addplot [mark=asterisk, violet!80, dashed, 
                      mark options={solid},
                      error bars/.cd, y dir=both, y explicit,
                      error bar style={line width=1pt,solid},
                      error mark options={line width=1pt,mark size=2pt,rotate=90}]
                    table [x=x, y=y, y error=y-err]{
                        x	y y-err
                        0.3 	0.01427423123 0.005
                        0.2 	0.02439524751 0.01
                        0.1 	0.04294733276 0.017
                        0.05	0.06745922205 0.030
                            };
            \addplot [mark=o, magenta!80,
                      mark options={solid},
                      error bars/.cd, y dir=both, y explicit,
                      error bar style={line width=1pt,solid},
                      error mark options={line width=1pt,mark size=2pt,rotate=90}]
                    table [x=x, y=y, y error=y-err]{
                        x	y y-err
                        0.3 	0.01883239322 0.008
                        0.2 	0.03442594473 0.015
                        0.1 	0.05805116958 0.024
                        0.05	0.07859456268 0.04
                            };
            \addplot [mark=asterisk, red!80,
                      mark options={solid},
                      error bars/.cd, y dir=both, y explicit,
                      error bar style={line width=1pt,solid},
                      error mark options={line width=1pt,mark size=2pt,rotate=90}]
                    table [x=x, y=y, y error=y-err]{
                        x	y y-err
                        0.3 	0.05256431142 0.02
                        0.2 	0.1062422903 0.046
                        0.1 	0.1158602629 0.06
                        0.05	0.1144569436 0.06
                            };
            \end{axis} 
        \end{tikzpicture}

%% file: figures/tikz_diagrams/qa_coord_dist.tex
\begin{tikzpicture}[scale=1.0] 
            \begin{axis}[legend pos=north east,  width={1.1\textwidth},
                ylabel=Dist Mean Causal Effect,
    xlabel=Counterfactual Loss,
    xmin=0.025, xmax=0.325,
    ymin=0.0,ymax=0.15,
                height=4.2cm,
    y tick label style={
        /pgf/number format/.cd,
        fixed,
        fixed zerofill,
        precision=2,
        /tikz/.cd
    }]
            \addplot [black!80, 
                      mark options={solid},
                      error bars/.cd, y dir=both, y explicit,
                      error bar style={line width=1pt,solid},
                      error mark options={line width=1pt,mark size=2pt,rotate=90}]
                    table [x=x, y=y, y error=y-err]{
                        x	y y-err
                        0.3	    0.003611974971 0.002
                        0.2	    0.006599238547 0.002
                        0.1	    0.01110863691 0.008
                        0.05    0.01591903733 0.0012
                            };
            \addplot [mark=o, blue!80, dashed, 
                      mark options={solid},
                      error bars/.cd, y dir=both, y explicit,
                      error bar style={line width=1pt,solid},
                      error mark options={line width=1pt,mark size=2pt,rotate=90}]
                    table [x=x, y=y, y error=y-err]{
                        x	y y-err
                        0.3 	0.007596768935 0.0005
                        0.2 	0.01248876882 0.0008
                        0.1 	0.02142388239 0.0013
                        0.05	0.03220937201 0.0021
                            };
            \addplot [mark=asterisk, violet!80, dashed, 
                      mark options={solid},
                      error bars/.cd, y dir=both, y explicit,
                      error bar style={line width=1pt,solid},
                      error mark options={line width=1pt,mark size=2pt,rotate=90}]
                    table [x=x, y=y, y error=y-err]{
                        x	y y-err
                        0.3 	0.01217582975 0.008
                        0.2 	0.02001205767 0.0015
                        0.1 	0.03681765557 0.0034
                        0.05	0.05433945985 0.0042
                            };
            \addplot [mark=o, magenta!80,
                      mark options={solid},
                      error bars/.cd, y dir=both, y explicit,
                      error bar style={line width=1pt,solid},
                      error mark options={line width=1pt,mark size=2pt,rotate=90}]
                    table [x=x, y=y, y error=y-err]{
                        x	y y-err
                        0.3 	0.0136969058 0.01
                        0.2 	0.02525038489 0.0018
                        0.1 	0.0511779239 0.0043
                        0.05	0.05418066253 0.05
                            };
            \addplot [mark=asterisk, red!80,
                      mark options={solid},
                      error bars/.cd, y dir=both, y explicit,
                      error bar style={line width=1pt,solid},
                      error mark options={line width=1pt,mark size=2pt,rotate=90}]
                    table [x=x, y=y, y error=y-err]{
                        x	y y-err
                        0.3 	0.02557342348 0.013
                        0.2 	0.05929853081 0.036
                        0.1 	0.06666958426 0.045
                        0.05	0.03285705811 0.062
                            };
            \end{axis} 
        \end{tikzpicture}

%% file: figures/tikz_diagrams/qa_npvp_dist.tex
\setlength{\fboxsep}{0pt}
        \begin{tikzpicture}[scale=1.0] 
            \begin{axis}[legend pos=north east,  width={1.1\textwidth},
    xlabel=Counterfactual Loss,
    xmin=0.025, xmax=0.325,
    ymin=0.0,ymax=0.03,
                height=4.2cm,
    y tick label style={
        /pgf/number format/.cd,
        fixed,
        fixed zerofill,
        precision=1,
        /tikz/.cd
    }]
            \addplot [black!80, 
                      mark options={solid},
                      error bars/.cd, y dir=both, y explicit,
                      error bar style={line width=1pt,solid},
                      error mark options={line width=1pt,mark size=2pt,rotate=90}]
                    table [x=x, y=y, y error=y-err]{
                        x	y y-err
                        0.3	    0.0004235868744 0.0005
                        0.2	    0.0005816616051 0.0012
                        0.1	    0.0003914012215 0.0016
                        0.05    -0.000108033719 0.0021
                            };
            \addplot [mark=o, blue!80, dashed, 
                      mark options={solid},
                      error bars/.cd, y dir=both, y explicit,
                      error bar style={line width=1pt,solid},
                      error mark options={line width=1pt,mark size=2pt,rotate=90}]
                    table [x=x, y=y, y error=y-err]{
                        x	y y-err
                        0.3 	0.0009391645244 0.0001
                        0.2 	0.002066930641 0.00016
                        0.1 	0.003712925127 0.0023
                        0.05	0.005042009836 0.0034
                            };
            \addplot [mark=asterisk, violet!80, dashed, 
                      mark options={solid},
                      error bars/.cd, y dir=both, y explicit,
                      error bar style={line width=1pt,solid},
                      error mark options={line width=1pt,mark size=2pt,rotate=90}]
                    table [x=x, y=y, y error=y-err]{
                        x	y y-err
                        0.3 	0.0008174671637 0.0013
                        0.2 	0.001776454711 0.0016
                        0.1 	0.002466292207 0.0024
                        0.05	0.002361193803 0.004
                            };
            \addplot [mark=o, magenta!80,
                      mark options={solid},
                      error bars/.cd, y dir=both, y explicit,
                      error bar style={line width=1pt,solid},
                      error mark options={line width=1pt,mark size=2pt,rotate=90}]
                    table [x=x, y=y, y error=y-err]{
                        x	y y-err
                        0.3 	0.001334184357 0.0017
                        0.2 	0.002384578928 0.0026
                        0.1 	0.001026776904 0.0032
                        0.05	-0.0002235446297 0.006
                            };
            \addplot [mark=asterisk, red!80,
                      mark options={solid},
                      error bars/.cd, y dir=both, y explicit,
                      error bar style={line width=1pt,solid},
                      error mark options={line width=1pt,mark size=2pt,rotate=90}]
                    table [x=x, y=y, y error=y-err]{
                        x	y y-err
                        0.3 	0.0007731357388 0.002
                        0.2 	0.001356856507 0.0032
                        0.1 	0.001215714298 0.006
                        0.05	-0.001427125468 0.010
                            };
            \end{axis} 
        \end{tikzpicture}

%% file: figures/tikz_diagrams/qa_rc_dist.tex
\setlength{\fboxsep}{0pt}
\begin{tikzpicture}[scale=1.0] 
            \begin{axis}[legend pos=north east,  width={1.1\textwidth},
    xlabel=Counterfactual Loss,
    xmin=0.025, xmax=0.325,
    ymin=0.0,ymax=0.03,
  legend style={fill=none, draw=none},
                height=4.2cm,
    y tick label style={
        /pgf/number format/.cd,
        fixed,
        fixed zerofill,
        precision=1,
        /tikz/.cd
    }]
            \addplot [black!80, 
                      mark options={solid},
                      error bars/.cd, y dir=both, y explicit,
                      error bar style={line width=1pt,solid},
                      error mark options={line width=1pt,mark size=2pt,rotate=90}]
                    table [x=x, y=y, y error=y-err]{
                        x	y y-err
                        0.3	    0.000381131093 0.001
                        0.2	    0.0009474364662 0.002
                        0.1	    0.00277550603 0.0003
                        0.05    0.00307942681 0.00042
                            };
            \addplot [mark=o, blue!80, dashed, 
                      mark options={solid},
                      error bars/.cd, y dir=both, y explicit,
                      error bar style={line width=1pt,solid},
                      error mark options={line width=1pt,mark size=2pt,rotate=90}]
                    table [x=x, y=y, y error=y-err]{
                        x	y y-err
                        0.3 	0.0009923713928 0.002
                        0.2 	0.002045414032 0.0031
                        0.1 	0.003102062694 0.0006
                        0.05	0.004345328026 0.0007
                            };
            \addplot [mark=asterisk, violet!80, dashed, 
                      mark options={solid},
                      error bars/.cd, y dir=both, y explicit,
                      error bar style={line width=1pt,solid},
                      error mark options={line width=1pt,mark size=2pt,rotate=90}]
                    table [x=x, y=y, y error=y-err]{
                        x	y y-err
                        0.3 	0.002710274976 0.0003
                        0.2 	0.005305175561 0.005
                        0.1 	0.007528469463 0.007
                        0.05	0.01056693335 0.0012
                            };
            \addplot [mark=o, magenta!80,
                      mark options={solid},
                      error bars/.cd, y dir=both, y explicit,
                      error bar style={line width=1pt,solid},
                      error mark options={line width=1pt,mark size=2pt,rotate=90}]
                    table [x=x, y=y, y error=y-err]{
                        x	y y-err
                        0.3 	0.006558840305 0.005
                        0.2 	0.008510509678 0.008
                        0.1 	0.01312092671 0.016
                        0.05	0.02196327241 0.021
                            };
            \addplot [mark=asterisk, red!80,
                      mark options={solid},
                      error bars/.cd, y dir=both, y explicit,
                      error bar style={line width=1pt,solid},
                      error mark options={line width=1pt,mark size=2pt,rotate=90}]
                    table [x=x, y=y, y error=y-err]{
                        x	y y-err
                        0.3 	0.003268094458 0.01
                        0.2 	0.0128650812 0.018
                        0.1 	0.004868534351 0.020
                        0.05	0.01359376148 0.03
                            };
            \end{axis} 
        \end{tikzpicture}